\documentclass{article} 
\usepackage{iclr2025_conference,times}

\usepackage{amsmath,amsfonts,bm}

\def\eqref#1{equation~\ref{#1}}

\def\1{\bm{1}}

\DeclareMathAlphabet{\mathsfit}{\encodingdefault}{\sfdefault}{m}{sl}
\SetMathAlphabet{\mathsfit}{bold}{\encodingdefault}{\sfdefault}{bx}{n}

\usepackage{url}
\usepackage{xspace}

\usepackage[utf8]{inputenc} 
\usepackage[T1]{fontenc}    
\usepackage{booktabs}       
\usepackage{amsfonts}       
\usepackage{nicefrac}       
\usepackage{microtype}      
\usepackage{xcolor}         
\definecolor{MegaScience}{HTML}{1d4ed8}

\usepackage{hyperref}
\hypersetup{
    colorlinks=true,
    citecolor=MegaScience,
    linkcolor=MegaScience,   
    urlcolor=MegaScience      
}

\usepackage{fancyhdr}
\usepackage{ifthen}

\usepackage{graphicx}
\usepackage{tabularx}
\usepackage{lscape}
\usepackage{array}
\usepackage{tikz}
\usepackage{placeins}
\usepackage{tcolorbox}
\usepackage{amsmath}
\usepackage{subcaption}
\usepackage{listings}
\usepackage{amsfonts}
\usepackage{lipsum}
\usepackage{pifont}

\usepackage{pgfplots}
\usepackage{pgfplotstable}
\pgfplotsset{compat=1.17}
\usepackage{wrapfig}
\usepackage{multirow}
\usepackage{colortbl}
\usepackage{enumitem}
\usepackage{fancyvrb}
\usepackage{fvextra}
\usepackage{makecell}
\usepackage{wrapfig}

\usepackage{amssymb}
\usepackage{pifont}

\usepackage{newunicodechar}
\newunicodechar{π}{\ensuremath{\pi}}

\usepackage{diagbox}

\usepackage{caption}

\definecolor{Math}{HTML}{1E3A8A}
\definecolor{Physics}{HTML}{BB9727}
\definecolor{Chemistry}{HTML}{54B345}
\definecolor{Biology}{HTML}{32B897}
\definecolor{Geography}{HTML}{05B9E2}
\definecolor{Astronomy}{HTML}{8983BF}
\definecolor{Computer Science}{HTML}{C76DA2}

\definecolor{easy}{HTML}{8ECFC9}
\definecolor{medium}{HTML}{FFBE7A}
\definecolor{hard}{HTML}{FA7F6F}

\definecolor{rule}{HTML}{F27970}
\definecolor{model}{HTML}{BB9727}
\definecolor{answer}{HTML}{54B345}
\definecolor{process}{HTML}{05B9E2}

\definecolor{lightblue}{RGB}{173, 216, 230}
\definecolor{mybrown}{RGB}{128,64,0}
\gdef\Sepline{%
  \par\noindent\makebox[\linewidth][l]{%
  \hspace*{-\mdflength{innerleftmargin}}%
   \tikz\draw[thick,dashed,gray!60] (0,0) --%
        (\textwidth+\the\mdflength{innerleftmargin}+\the\mdflength{innerrightmargin},0);
  }\par\nobreak}

\definecolor{wkblue}{RGB}{179,229,226}
\definecolor{meta-color}{RGB}{16,177,168}

\title{\includegraphics[height=1.5em]{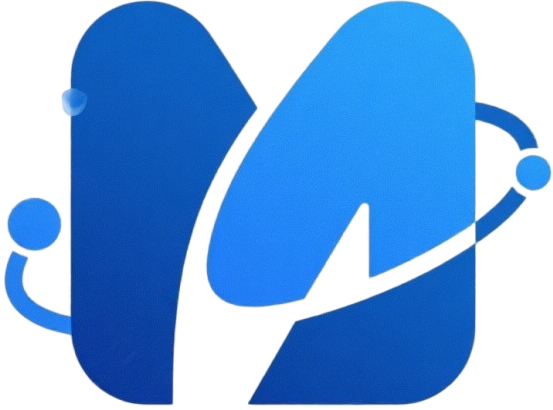}\textcolor{MegaScience}{MegaScience}: Pushing the Frontiers of Post-Training Datasets for Science Reasoning}

\definecolor{DeepRed}{HTML}{B22222}
\newcommand{\cofirst}{\textcolor{MegaScience}{$^{\text{\ding{170}}}$}}
\newcommand{\corresp}{\textcolor{DeepRed}{$^{\spadesuit}$}}

\newcommand{\github}{\raisebox{-1.5pt}{\includegraphics[height=1.05em]{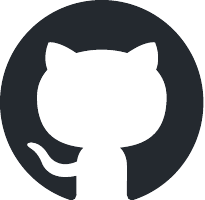}}\xspace}
\newcommand{\huggingface}{\raisebox{-1.5pt}{\includegraphics[height=1.05em]{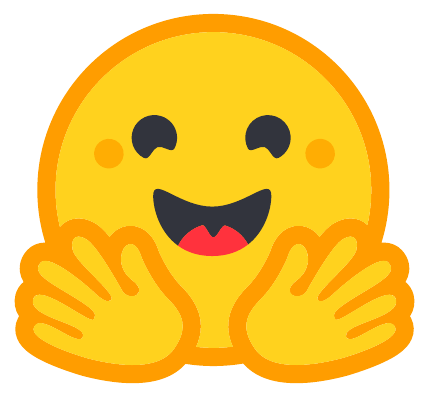}}\xspace}

\author{Run-Ze Fan\cofirst, \space\space Zengzhi Wang\cofirst, \space\space Pengfei Liu\corresp \\
Shanghai Jiao Tong University, \space\space SII, \space\space GAIR Lab \\
\texttt{runze.fan@icloud.com} \quad \texttt{\{zengzhi.wang, pengfei\}@sjtu.edu.cn}\\
    \\
    \github \href{https://github.com/GAIR-NLP/MegaScience}{\textbf{GAIR-NLP/MegaScience}} ~ ~ ~ \huggingface \href{https://huggingface.co/MegaScience}{\textbf{MegaScience}}  ~ ~ ~ \github \href{https://github.com/GAIR-NLP/lm-open-science-evaluation}{\textbf{MegaScience-Eval}} \\
    \vspace{-5mm}
}

\newcommand{\ourdataset}{\textsc{TextbookReasoning}\xspace}
\newcommand{\megascience}{\textsc{MegaScience}\xspace}
\newcommand{\ourname}{\textsc{MegaScience}\xspace}

\iclrfinalcopy 

\begin{document}

\maketitle



\thispagestyle{fancy}
\fancyhead{}
\lhead{\includegraphics[height=0.90cm]{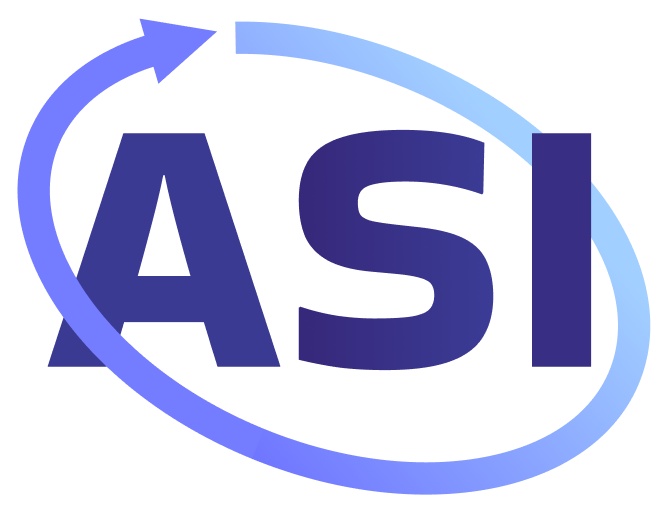}}
\rhead{%
  \raisebox{-0.1cm}{\includegraphics[height=0.7cm]{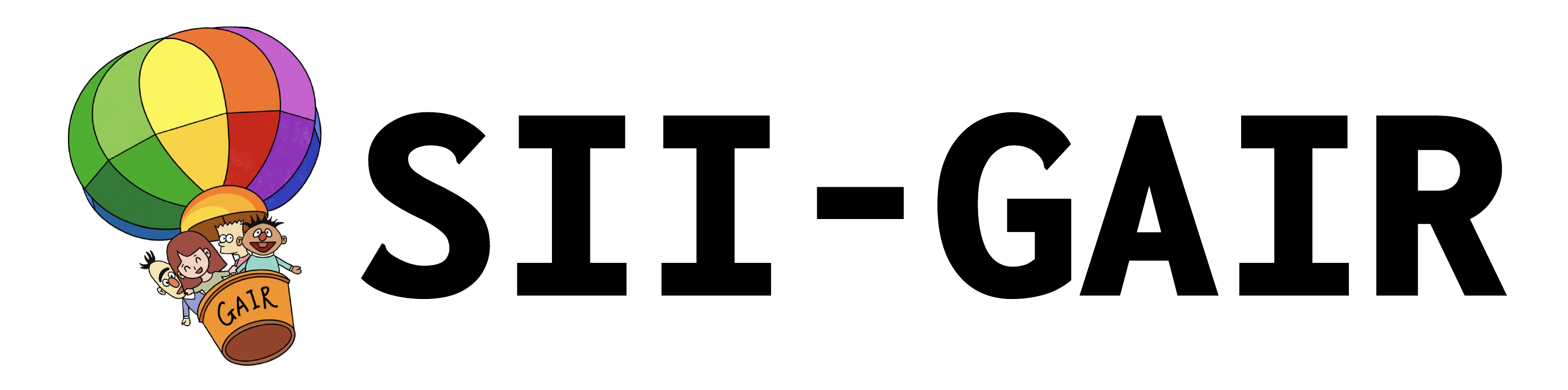}}%
}

\fancyfoot[C]{\thepage}

\fancypagestyle{plain}{
    \fancyhf{}
    \fancyhead[L]{\includegraphics[height=1cm]{figures/GAIR_Logo3.pdf}}
    \fancyhead[C]{}
    \fancyfoot[C]{\thepage}
    \renewcommand{\headrulewidth}{0pt}
}

\renewcommand{\thefootnote}{\fnsymbol{footnote}}
\footnotetext{\cofirst Equal contribution.\quad\corresp Corresponding author.}

\renewcommand{\thefootnote}{\arabic{footnote}}
\begin{abstract}Scientific reasoning is critical for developing AI scientists and supporting human researchers in advancing the frontiers of natural science discovery. However, the open-source community has primarily focused on mathematics and coding while neglecting the scientific domain, largely due to the absence of open, large-scale, high-quality, verifiable scientific reasoning datasets. To bridge this gap, we first present \textbf{\ourdataset}, an open dataset featuring truthful reference answers extracted from 12k university-level scientific textbooks, comprising 650k reasoning questions spanning 7 scientific disciplines. We further introduce \textbf{\megascience}, a large-scale mixture of high-quality open-source datasets totaling 1.25 million instances, developed through systematic ablation studies that evaluate various data selection methodologies to identify the optimal subset for each publicly available scientific dataset. Meanwhile, we build a comprehensive evaluation system covering diverse subjects and question types across 15 benchmarks, incorporating comprehensive answer extraction strategies to ensure accurate evaluation metrics. Our experiments demonstrate that our datasets achieve superior performance and training efficiency with more concise response lengths compared to existing open-source scientific datasets. Furthermore, we train Llama3.1, Qwen2.5, and Qwen3 series base models on \megascience, which significantly outperform the corresponding official instruct models in average performance. In addition, \textbf{\megascience exhibits greater effectiveness for larger and stronger models, suggesting a scaling benefit for scientific tuning}. We release our data curation pipeline, evaluation system, datasets, and seven trained models to the community to advance scientific reasoning research.
\end{abstract}

\vspace{7pt}

\begin{figure}[htbp]
    \centering
    \begin{minipage}{0.4\textwidth}
        \centering
        \includegraphics[width=\textwidth]{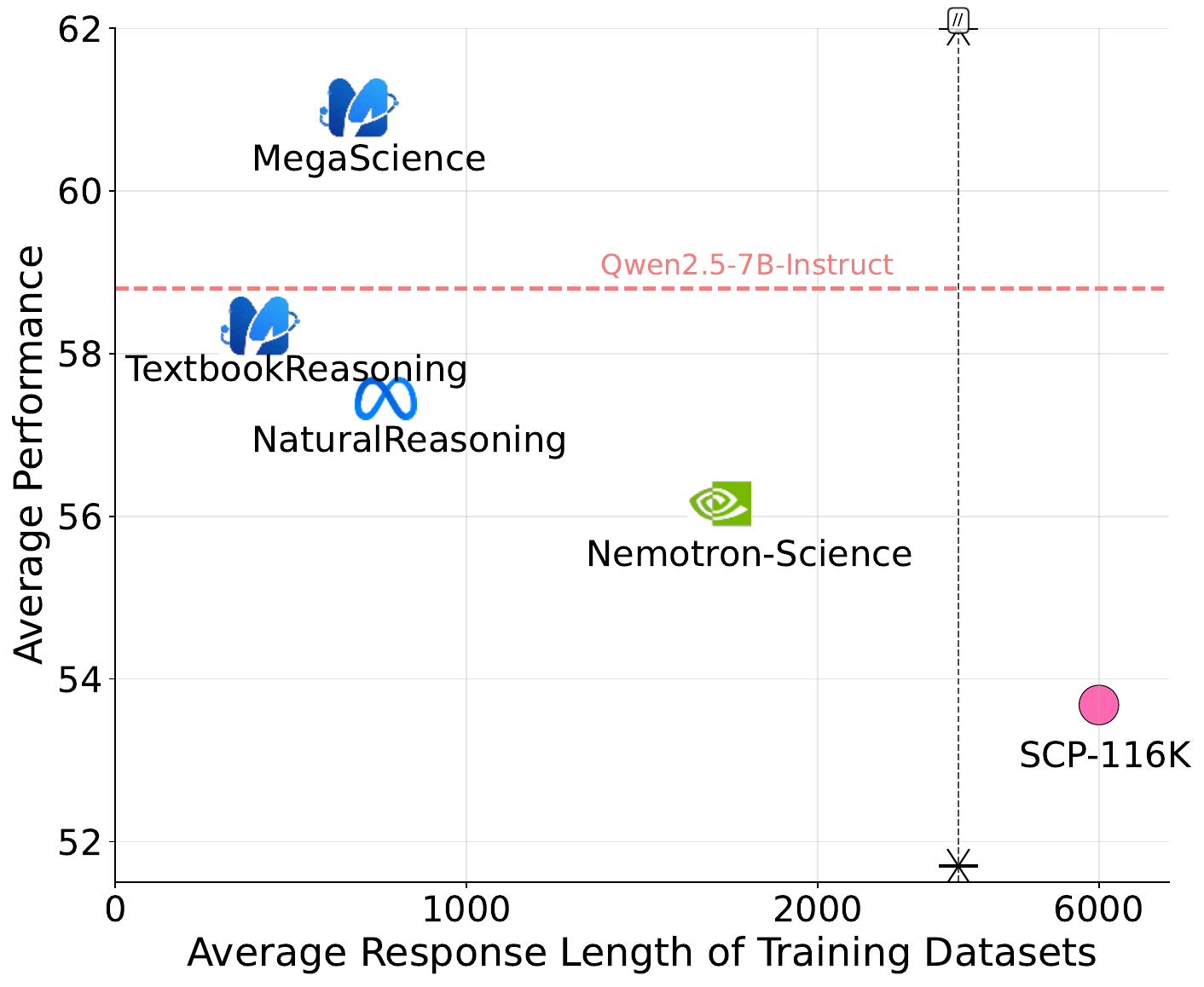}
        \caption{Trade-off between model performance and inference efficiency (average response length) on Qwen2.5-7B.}
        \label{fig:response_length_vs_performance}
    \end{minipage}
    \hfill
    \begin{minipage}{0.58\textwidth}
        \centering
        \includegraphics[width=\textwidth]{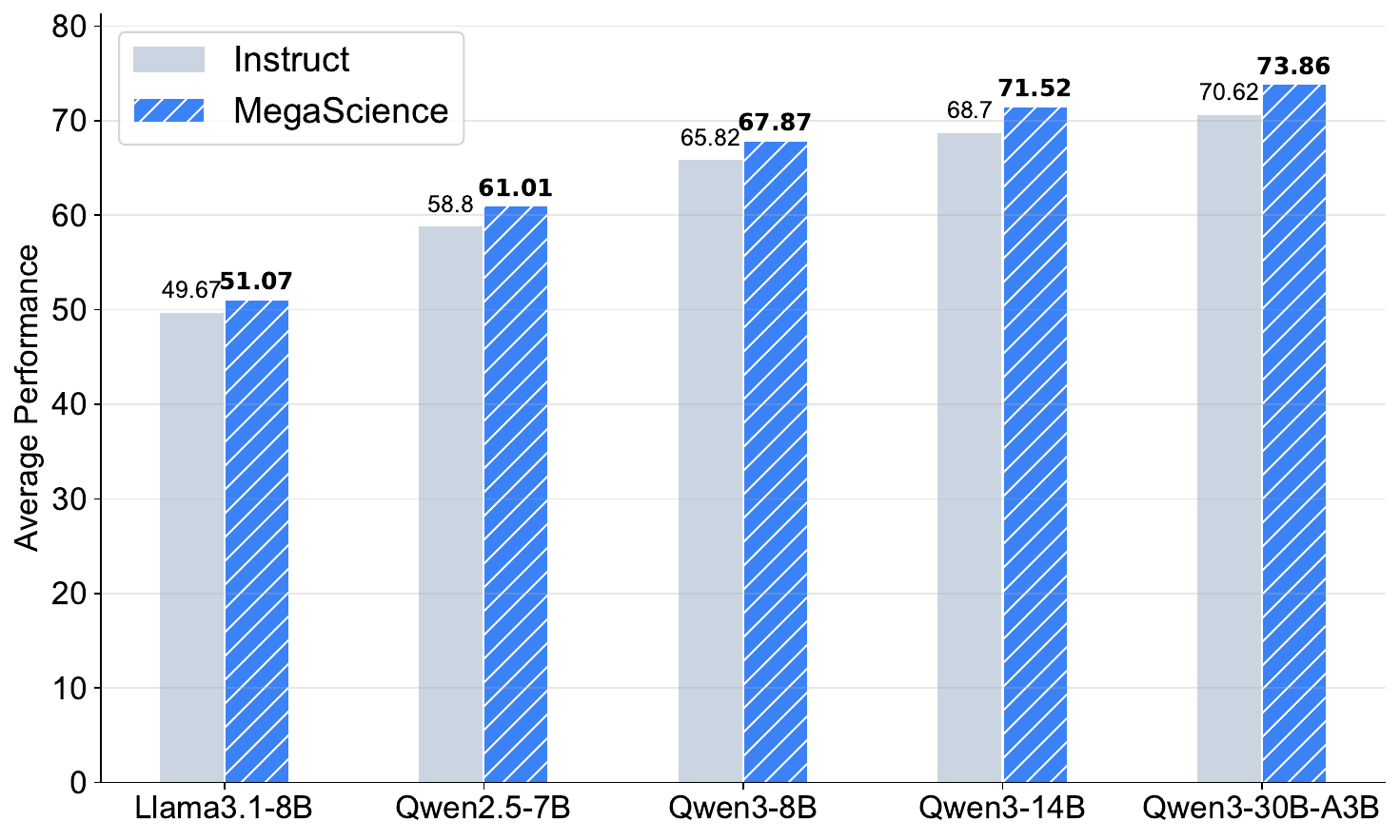}
        \caption{Comparison of base models trained on \megascience vs. official instruct models (non-thinking).}
        \label{fig:comparision_megascience_instruct}
    \end{minipage}
\end{figure}

\clearpage
\begin{figure}[t]
    \centering
    \includegraphics[width=0.8\textwidth]{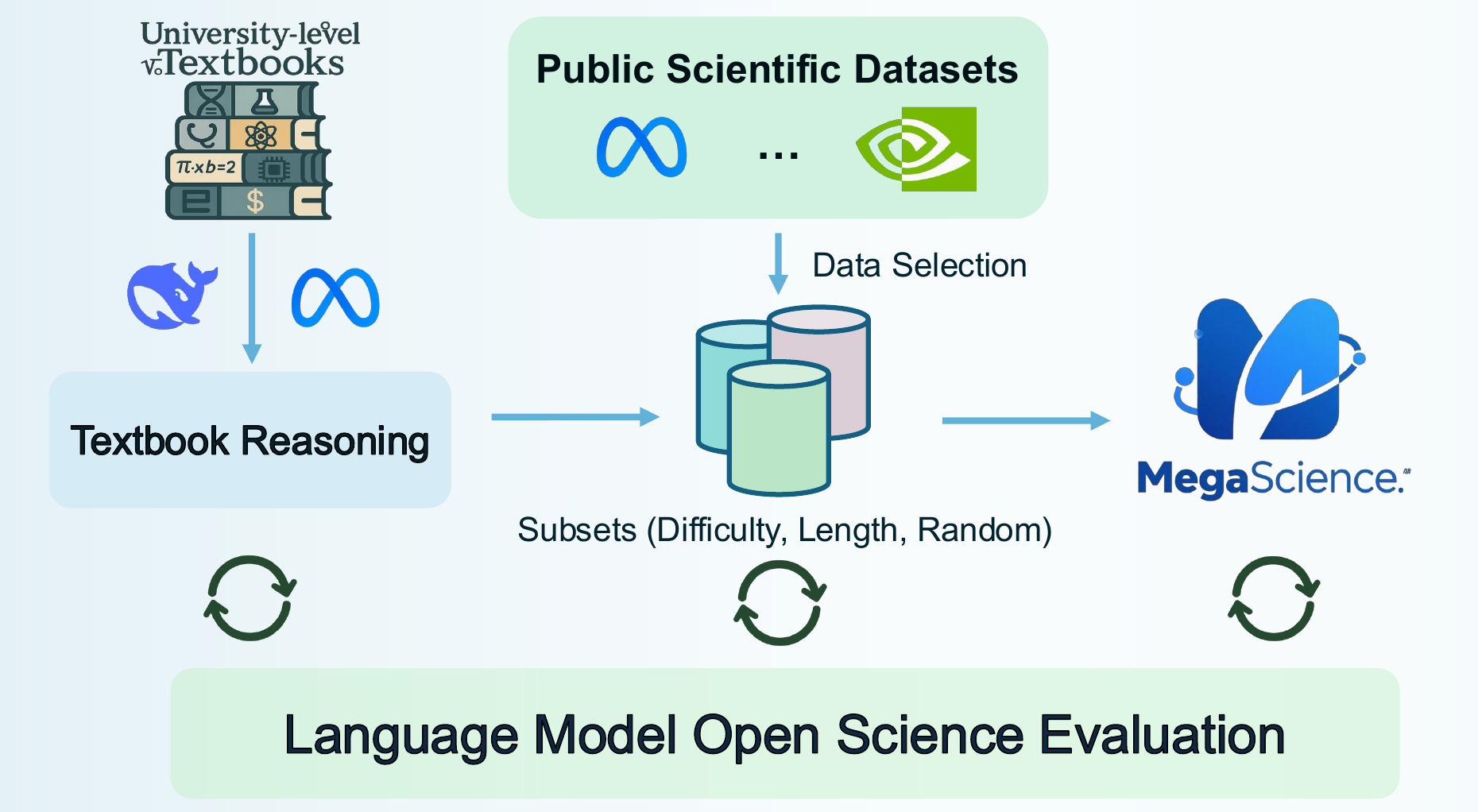}
    \caption{The overall of \textsc{MegaScience} datasets.}
    \label{fig:mixture_figure}
\end{figure}
\section{Introduction}


Large Language Models (LLMs) have evolved from knowledge retrieval systems into cognitive reasoning systems~\citep{xia2025generativeaiactii}, representing a significant milestone toward Artificial General Intelligence (AGI)~\citep{jaech2024openaio1, guo2025deepseekr1}. These reasoning models have primarily focused on mathematics and coding, as these domains provide abundant datasets, established benchmarks, and well-defined verification mechanisms~\citep{zhou2025megamath, tsoukalas2024putnambench, liu2024acemath, wang2024mathpile, jimenez2023swe}. Scientific reasoning represents another critical capability that is essential for developing AI scientists and assisting human researchers in advancing the frontiers of natural science~\citep{jumper2021alphafold, yang2023alphafold2}. However, scientific reasoning remains significantly underdeveloped compared to mathematics and coding, particularly within the open-source community.

Despite the availability of some open-source scientific reasoning datasets, several critical challenges remain unaddressed: 

(1) \textbf{Unreliable benchmark evaluation}: Many open-source scientific benchmarks adopt multiple-choice formats, which, while easy to implement, oversimplify the complexity of scientific reasoning. Consequently, post-training datasets in scientific domains often follow this format to maintain distributional consistency (e.g., Nemotron-Science~\citep{bercovich2025nemotron}). However, our observations reveal that models trained on such data exhibit inflated performance on multiple-choice evaluations but struggle significantly with computational tasks, suggesting a disconnect between benchmark performance and true reasoning ability. 

(2) \textbf{Less rigorous decontamination}: Existing decontamination techniques typically rely on n-gram overlap or embedding similarity to remove potential benchmark leakage. These methods are inherently fragile, easily circumvented by minor variations in phrasing or structure, and thus fail to ensure the integrity of benchmark evaluations. We found substantial overlap with benchmarks from most existing post-training datasets on science domains. 

(3) \textbf{Low-quality reference answers}: Reference answers in many scientific datasets are either scraped from web sources (e.g., NaturalReasoning~\citep{yuan2025naturalreasoning}) or generated by LLMs (e.g., Nemotron-Science~\citep{bercovich2025nemotron}). Both methods suffer from increasing unreliability—web content is now saturated with AI-generated text, and LLMs themselves are prone to hallucination—making it difficult to guarantee the factual accuracy and scientific rigor of the answers.

(4) \textbf{Superficial knowledge (data) distillation}:  A common practice involves distilling data from large reasoning models—such as directly prompting DeepSeek-R1~\citep{guo2025deepseekr1} to generate long chain of thoughts (CoT)~\citep{wei2022cot} solutions (e.g., NaturalThoughts~\citep{li2025naturalthoughts} and Nemotron-Science~\citep{bercovich2025nemotron}). While intuitive and easy to implement, it remains largely superficial. The resulting CoT data are often prone to overthinking~\citep{chen2024overthink}, which also brings challenges in training especially for small models and inference efficiency. Such shallow operations hinder the more principled, efficient, and generalizable knowledge transfer.

To bridge this gap, we first introduce \textbf{\ourdataset} (\S\ref{sec:textbookreasoning_data_curation}), an open-source university-level scientific post-training dataset with truthful reference answers, extracted from nearly 12k university-level scientific textbooks, comprising 650k reasoning questions spanning various topics, including physics, biology, chemistry, medicine, computer science, mathematics, and economics. Specifically, our data curation pipeline consists of textbook digitalization, dual QA pairs extraction, deduplication, QA pairs refinement, filtering, and LLM-based decontamination. This pipeline, fully automated through LLMs, facilitates the scalable acquisition of high-quality datasets. 

To further advance open-source post-training datasets for scientific reasoning, we introduce \textbf{\megascience} (\S\ref{sec:mixture_data_curation}), a large-scale mixture of high-quality open-source datasets consisting of 1.25 million instances. We first collect multiple public datasets, then conduct comprehensive ablation studies across different data selection methods to identify the optimal approach for each dataset, thereby contributing high-quality subsets. Furthermore, we annotate step-by-step solutions for all datasets except \ourdataset. 

To facilitate scientific reasoning development in the open-source community, we design and open-source an evaluation framework (\S\ref{sec:evaluation_framework}) covering diverse subjects (e.g., biology and physics) and question types (e.g., multiple-choice questions and computational problems) across 15 benchmarks. This framework enables easy reproduction of our experimental results and fair comparison across different models by providing equitable treatment. Additionally, we design comprehensive answer extraction strategies to ensure the accuracy of final evaluation metrics. 

Our supervised fine-tuning experiments (\S\ref{sec:supervised_finetuning}) demonstrate that our datasets not only enable efficient training and inference but also achieve state-of-the-art performance in the scientific domain. Finally, we train Llama3.1, Qwen2.5, and Qwen3 series base models on \megascience, which outperform the official instruct models in average performance, successfully advancing the frontiers of the open-source community in the science domain. We find that \megascience exhibits greater effectiveness for larger and stronger models, suggesting a scaling benefit for scientific instruction tuning.

Our contribution can be summarized as follows:

\begin{enumerate}[leftmargin=20pt, label=(\arabic*)]
    \item We present \ourdataset and \megascience, two datasets that advance the frontier in the scientific domain by enabling base models to outperform official instruct models on scientific tasks when fine-tuned with our data. In addition, \megascience exhibits greater effectiveness for larger and stronger models, suggesting a scaling benefit for scientific tuning.
    \item Our datasets contain shorter responses (410 tokens for \ourdataset and 721 for \megascience), which not only make training and inference efficient but also achieve state-of-the-art performance in the scientific domain.
    \item We release our data curation pipeline, evaluation system, datasets, and trained models to the community to advance scientific reasoning research.
\end{enumerate}

\section{\ourdataset Data Curation}
\label{sec:textbookreasoning_data_curation}
\begin{figure}[t]
    \centering
    \includegraphics[width=\textwidth]{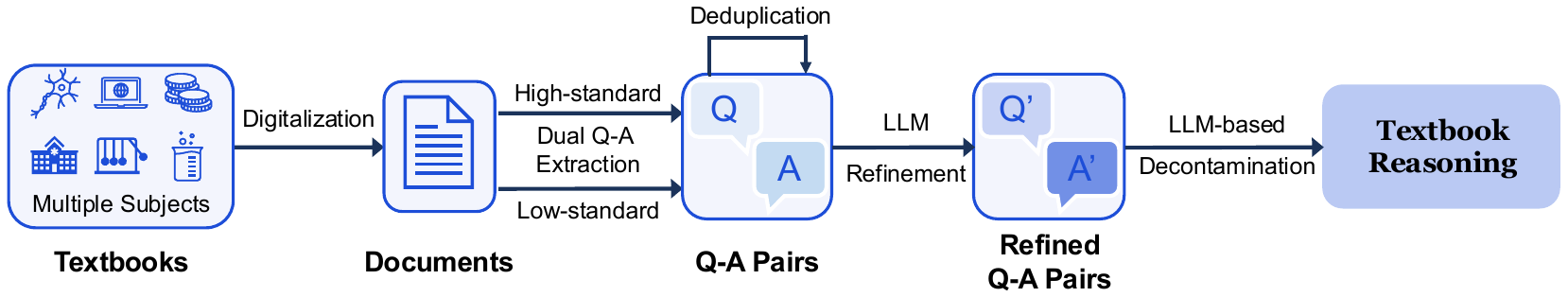}
    \caption{The pipeline of \ourdataset data curation.}
    \label{fig:textbook_reasoning_figure}
\end{figure}

Current scientific datasets are predominantly derived from web sources or generated through LLM distillation, resulting in a lack of large-scale, challenging, and diverse questions accompanied by truthful reference answers. Textbooks serve as naturally reliable sources of information, as they are meticulously crafted by human experts and embody accumulated human knowledge. Moreover, textbooks offer a more systematic and coherent knowledge structure than web data, which makes them better suited for knowledge learning in LLMs. The superiority of such human-curated content has been demonstrated in serious works on phi models~\citep{gunasekar2023phi1, li2023phi1.5} during pretraining, which show that textbooks exhibit significantly higher information density than web data. However, existing research has not yet explored how to effectively leverage textbooks for developing scientific reasoning capabilities in LLMs during post-training. To address this gap, we propose a comprehensive pipeline designed to maximize the educational value extracted from textbooks. This pipeline introduces \ourdataset, an open-source university-level scientific post-training dataset featuring verified reference answers. The dataset is derived from 12.8k university-level scientific textbooks and comprises 651k reasoning questions spanning diverse disciplines, including physics, biology, chemistry, medicine, computer science, mathematics, and economics. An overview of the data curation pipeline is illustrated in Figure \ref{fig:textbook_reasoning_figure}.

\subsection{Textbooks Collection and Digitization}
We collected a large corpus of books by crawling PDF documents from the web. To address copyright concerns, we filtered out books marked as restricted for public access based on their metadata information. Subsequently, we employed Llama3.3-70B-Instruct~\citep{grattafiori2024llama3} to automatically classify each book's subject area and academic level, excluding materials below university level to ensure appropriate difficulty. This filtering process yielded a final dataset comprising 12.8k academic books across seven disciplines: 2,305 books in medicine and biology, 1,017 books in chemistry, 6,057 books in computer science and artificial intelligence, 1,685 books in physics, 1,578 books in mathematics, and 158 books in economics. Finally, we employ olmOCR~\citep{poznanski2025olmocr}~\footnote{\href{https://olmocr.allenai.org/}{https://olmocr.allenai.org/}} to convert PDF documents into machine-readable text.


\begin{table}[htbp]
    \caption{Q-A Extraction Statistics}
    \centering
    \resizebox{\textwidth}{!}{ 
    \begin{tabular}{l|ccccc}
    \toprule
    \textbf{Subject} & \textbf{\# Books} & \textbf{\# Chunks} & \textbf{\# Valid Chunks} & \textbf{\# Extracted Pairs (High)} & \textbf{\# Extracted Pairs (Low)} \\
    \midrule
    Biology & 2,305 & 119,581 & 6,929 & 1,394 & 102,926 \\
    Chemistry & 1,017 & 49,847 & 5,490   & 1,979 & 70,756 \\
    Computer Science & 6,057 & 116,380 & 5,521 & 5,890 & 16,322 \\
    Economics & 158 & 8,071 & 329 & 94 & 1,851\\
    Mathematics & 1578 & 56,952 & 35,876 & 6,376 & 553,786 \\
    Medicine & 2,305 & 119,581 & 9,797 & 4,919 & 120,296\\
    Physics & 1,685 & 75,722 & 8,606 & 4,831 & 54,263 \\
    \midrule
    \rowcolor{MegaScience!20}
    Total & 12,800 & 546,134 & 72,548 & 25,483 & 920,200 \\
    \bottomrule
    \end{tabular}
    }
    \label{tab:qa_extraction}
\end{table}
\subsection{Dual Q-A Pairs Extraction}
Compared to question synthesis from given documents~\citep{li2023self}, Q-A pair extraction preserves more original information without introducing substantial LLM-generated content and avoids many conceptual questions such as ``what is'' queries. Unlike existing extraction pipelines, which only employ a single standard to extract questions~\citep{yue2024mammoth2}, we design a dual-extraction strategy with both high-standard and low-standard criteria to comprehensively mine complete Q-A pairs from the text, ensuring we capture content across varying levels of clarity and structure. Specifically, we segmented textbooks into 4,096-token chunks and processed each chunk through Llama3.3-70B-Instruct to extract Q-A pairs using two distinct criteria (refer to \ref{appx_prompts_extraction} for the detailed prompts). The high-standard criterion requires that questions demand multi-step reasoning rather than simple definition or concept recall, and that source documents contain comprehensive solutions with all necessary procedural steps. In contrast, the low-standard criterion requires only complete questions and answers. Table~\ref{tab:qa_extraction} presents the extraction statistics for each subject. We found substantial variations in the proportion of chunks containing questions across different disciplines. Mathematics exhibited the highest proportion of valid chunks, exceeding 60\%, whereas other disciplines demonstrated significantly lower rates, with fewer than 10\% of chunks containing questions. Finally, we acquire 945k extracted Q-A pairs.

\subsection{Question Deduplication}
\label{sec:q_deduplicatio}
To eliminate redundant questions from our dataset, we implement locality-sensitive min-hashing techniques~\footnote{\href{https://github.com/ChenghaoMou/text-dedup}{https://github.com/ChenghaoMou/text-dedup}} that operate at the word level. Questions exhibiting high similarity—defined by a threshold of 0.6—are systematically removed to prevent the inclusion of multiple variants that target identical reasoning tasks despite variations in their textual presentation.

\subsection{Q-A pair Refinement}
We find that many extracted questions may lack necessary information or contain citations to document information, while their corresponding answers often provide insufficient explanations and omit crucial intermediate reasoning steps. To address these issues, we employ DeepSeek-V3~\citep{liu2024deepseekv3} to refine the extracted Q-A pairs given the relevant source documents (see Figure~\ref{prompt:qa_refinement} for the prompt). The LLM ensures that refined questions incorporate all necessary contextual information and that refined answers provide comprehensive explanations with clear reasoning processes. Additionally, we use Llama3.3-70B-Instruct to identify question-answer pairs that lack reasoning processes (see Figure~\ref{prompt:identify_no_cot_qa} for prompt), and subsequently apply DeepSeek-V3 to add explanations and reformat the answers~\citep{fan-etal-2024-reformatted}.

After refinement, some questions still reference external sources, while others contain answers with contradictory reasoning, missing information, or invalid responses. We use Llama3.3-70B-Instruct to filter out these defective Q-A pairs (see Figure~\ref{prompt:filter_defective_qa} for the prompt).


\begin{table}[t]
    \caption{The numerical changes during \ourdataset curation.}
    \centering
    \resizebox{\textwidth}{!}{ 
    \begin{tabular}{l|ccccccc|c}
    \toprule
    \textbf{Actions} & \textbf{Biology} & \textbf{Chemistry} & \textbf{CS} & \textbf{Economics} & \textbf{Mathematics} & \textbf{Medicine} & \textbf{Physics} & \textbf{Total}\\
    \midrule
    Q-A Pairs & 104,320 & 72,735 & 22,212 & 1,945 & 560,162 & 125,215 & 59,094 & 945,683\\
    + Deduplication & 71,693 & 39,984 & 19,433 & 1,790 & 472,740 & 111,930 & 50,323 & 767,893\\
    + Filtering & 70,102 & 37,890 & 18,843 & 1,725 & 444,126 & 109,192 & 46,889 & 728,767 \\
    \rowcolor{MegaScience!20}
    + Decontamination & 52,850 & 32,157 & 17,742 & 1,296 & 424,714 & 81,638 & 41,443 & 651,840\\
    \bottomrule
    \end{tabular}
    }
    \label{tab:data_number_statistics}
\end{table}
\subsection{LLM-based Question Decontamination}
\label{sec:benchmark_decontamination}
Incorporating benchmark questions renders evaluation results unreliable~\citep{xu2024benchmarking, sainz-etal-2024-data}. To mitigate benchmark contamination, we examine potential overlap between \ourdataset and widely-used downstream benchmarks for evaluating LLMs' scientific reasoning capabilities, including MMLU~\citep{hendrycks2020mmlu}, GPQA~\citep{rein2024gpqa}, MMLU-Pro~\citep{wang2024mmlupro}, SuperGPQA~\citep{du2025supergpqa}, SciBench~\citep{wang2023scibench}, OlympicArena~\citep{huang2024olympicarena}, ChemBench~\citep{mirza2024chembench}, CS-Bench~\citep{song2024csbench}, MedQA~\citep{jin2020medqa}, MedMCQA~\citep{pal2022medmcqa}, PubMedQA~\citep{jin2019pubmedqa}, GSM8K~\citep{cobbe2021gsm8k}, and MATH~\citep{hendrycks2021math}. Traditional methods such as $n$-gram overlap are vulnerable to simple variations in test data (e.g., paraphrasing, translation), enabling rephrased samples to easily circumvent these basic detection techniques. To implement rigorous benchmark decontamination, we follow the approach of \citet{toshniwal2024openmathinstruct} and \citet{he2025deepmath} by deploying LLM-based decontamination through two main steps: (1) for each question, we use embedding similarity search (using BGE-large-en-v1.5~\citep{chen2024bge}) to identify the top-$k$ ($k=5$) most similar test examples from all benchmark datasets; (2) we create question pairs by matching each question with these top-$k$ test examples. Then, we deploy Llama3.3-70B-Instruct to evaluate whether any of these pairs constitute paraphrases via zero-shot prompting (see Figure~\ref{prompt:llm_judge_for_decontamination} for the prompt). If any of the $k$ pairs is determined to be a paraphrase, the question is removed from the dataset. The numerical changes for each step are presented in Table~\ref{tab:data_number_statistics}.

\section{\megascience Data Curation}
\label{sec:mixture_data_curation}

To further advance the frontiers of open-source post-training datasets for scientific reasoning, we collect multiple public datasets and explore different data selection methods and solution annotation techniques. Ultimately, we obtain a high-quality mixed dataset, \megascience, which consists of 1.25 million instances. An overall of the data recipe is illustrated in Figure~\ref{fig:mixture_figure}.

\subsection{Sourcing from Public Datasets}
We select NaturalReasoning~\citep{yuan2025naturalreasoning}, Nemotron-Science~\citep{bercovich2025nemotron}, and our \ourdataset as the source datasets. We exclude SCP-116K~\citep{lu2025scp} due to its inferior performance in scientific reasoning tasks.

\subsection{Question Deduplication and Decontamination}
We apply question deduplication and LLM-based question decontamination to NaturalReasoning and Nemotron-Science (details presented in \S\ref{sec:q_deduplicatio} and \S\ref{sec:benchmark_decontamination}).

\begin{figure}[t]
    \centering
    \includegraphics[width=\textwidth]{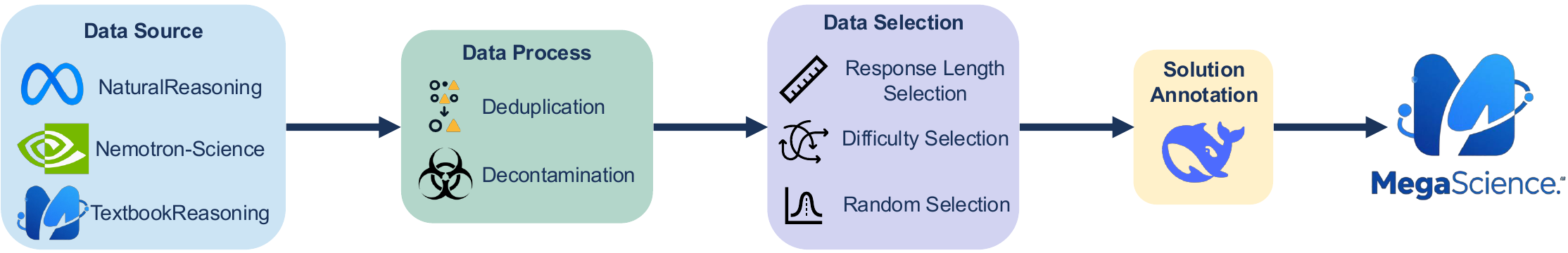}
    \caption{The overall of \megascience data recipe.}
    \label{fig:mixture_figure}
\end{figure}

\subsection{Data Selection}
\label{sec:data_selection}
Since indiscriminately mixing all available data would result in reduced training efficiency, we curate high-quality subsets from each dataset and combine these refined subsets for training. We design three data selection methods:
\begin{enumerate}[leftmargin=15pt, label=(\arabic*)]
    \item \textbf{Response Length Selection}: Following \citet{guha2025openthoughts}, which demonstrated that response length selection is the optimal method for the science domain, we annotate questions with Qwen2.5-72B-Instruct and retain the questions with the longest responses.
    \item \textbf{Difficulty Selection}: Since challenging questions are valuable for enhancing reasoning abilities, we design a difficulty selection method consisting of two steps: \textbf{(1) Reference answer annotation:} For \ourdataset, we employ Llama3.3-70B-Instruct to generate reference answers for each question-answer pair (see Figure~\ref{prompt:annotate_reference_answer} for the prompt). For NaturalReasoning, we directly use the provided reference answers. For Nemotron-Science, we utilize the summary portion of DeepSeek-R1's response as the reference answer. \textbf{(2) Difficulty evaluation:} To assess question difficulty, we follow the methodology of \citet{tong2024dart} by sampling 16 responses from Qwen2.5-7B-Instruct~\citep{yang2025qwen2.5} and using Qwen2.5-32B-Instruct to score each response on a scale of 0-10 relative to the reference answer (see Figure~\ref{prompt:evaluating_student_answer_with_reference} for the prompt). We then compute the average score across all sampled responses as the question's difficulty score, where a lower average score indicates higher difficulty. We filter out overly easy samples (average score $>$ 9) and potentially noisy samples (average score $<$ 1).
    \item \textbf{Random Selection}: Randomly select questions.
\end{enumerate}

\begin{table}[htbp]
    \caption{Performance comparison of data selection strategies. \textbf{General Avg.} denotes the average performance across general scientific reasoning tasks, \textbf{Specific Avg.} denotes the average performance across specific scientific reasoning tasks, and \textbf{Math Avg.} denotes the average performance across mathematical reasoning tasks (see \S\ref{sec:eval_suite} for details). \textbf{Bold} indicates the best results. \textcolor{MegaScience}{Blue} indicates the subset included in \megascience.}
    \centering
    \resizebox{\textwidth}{!}{ 
    \begin{tabular}{l|c|ccc|c}
    \toprule
    \textbf{Dataset}  & \textbf{Size (k)} & \textbf{General Avg.} & \textbf{Specific Avg.} & \textbf{Math Avg.} & \textbf{All Avg.} \\
    \midrule
    NaturalReasoning-DC & 1079 & 36.87 & \textbf{65.46} & \textbf{75.69} & \textbf{57.44} \\
    + Response Length Selection & 436.4 & \textbf{37.70} & 63.48 & 74.76 & 56.69 \\
    + Difficulty Selection & 436.4 & 	36.97 &	65.07 &	75.04 & 57.17 \\
    \rowcolor{MegaScience!20} 
    + Random Selection & 436.4 & 37.46 & 65.22 & 75.02 & 57.41 \\
    \midrule
    Nemotron-Science-DC & 447.4 & 35.16 & 67.56 & 68.33 & 56.15 \\
    + Response Length Selection & 173.3 & 34.33 & 67.43 &	71.09 & 56.39 \\
    \rowcolor{MegaScience!20} 
    + Difficulty Selection & 173.3 & \textbf{36.71} & \textbf{68.50} & \textbf{69.67} & \textbf{57.40} \\
    + Random Selection & 173.3 & 34.28 & 67.72 & 68.95 & 56.04\\
    \midrule
    \rowcolor{MegaScience!20} 
    \textsc{TextbookReasoning} & 651.8 & \textbf{39.58} & \textbf{65.15} & \textbf{75.93} & \textbf{58.33} \\
    + Response Length Selection & 297.6 & 36.94 & 62.53 &	75.57 & 56.18 \\
    + Difficulty Selection & 297.6 & 38.25 & 62.96 & 74.83 & 56.68 \\
    + Random Selection & 297.6 & 37.08 & 63.46 & 73.48 & 56.18\\
    \bottomrule
    \end{tabular}
    }
    \label{tab:data_selection_performance}
\end{table}

For each dataset, we first utilize difficulty selection to acquire $n$ instances, and then set the selection number for both response length selection and random selection to $n$ to ensure fair comparison. We choose the optimal data selection method for each dataset by conducting supervised fine-tuning on Qwen2.5-7B. The experimental results are shown in Table~\ref{tab:data_selection_performance}.

Random selection proves most effective for NaturalReasoning, while difficulty selection achieves optimal performance on Nemotron-Science. However, no single data selection method matches the performance of using the complete \ourdataset, suggesting it contains minimal low-quality instances. This finding supports retaining all instances in \megascience. The numerical changes for each step are detailed in Table~\ref{tab:mixture_data_statistics}.

\subsection{Solution Annotation}
\begin{wraptable}{r}{0.6\textwidth}
\caption{Statistics of the \megascience dataset. \textbf{Dedup} denotes question deduplication, \textbf{DC} represents LLM-based question decontamination, and \textbf{DS} indicates data selection.}
\centering
\small
\resizebox{0.6\textwidth}{!}{
\begin{tabular}{l|cccc}
    \toprule
    \textbf{Dataset} & \textbf{Raw Size} & \textbf{w/ Dedup} & \textbf{w/ DC} & \textbf{w/ DS} \\
    NaturalReasoning  & 1145.8k & 1145.8k & 1079k & 436.4k \\
    Nemotron-Science & 708.9k & 612k & 447.4k & 173.3k \\
    \textsc{TextbookReasoning} & 651.8k & 651.8k & 651.8k & 651.8k \\
    \midrule
    \rowcolor{MegaScience!20} 
    \megascience & 2506.5k & 2409.6k & 2178.2k & 1261.5k \\
    \bottomrule
    \end{tabular}}
    \label{tab:mixture_data_statistics}
\end{wraptable}
For \ourdataset, we retain the refined solution. For NaturalReasoning, we utilize DeepSeek-V3 to annotate step-by-step solutions due to the lower quality of the original responses generated by Llama3.3-70B-Instruct. For Nemotron-Science, DeepSeek-R1 generates excessively lengthy responses even for relatively simple questions~\citep{chen2024overthink}, which significantly reduces inference efficiency. To address this challenge, we utilize DeepSeek-V3 to annotate step-by-step solutions. To ensure data quality and conciseness, we filter out responses exceeding 4,096 tokens, as manual inspection reveals that overly long outputs often exhibit repetitive or redundant content. This step removes approximately 8,000 instances from the dataset.

\vspace{-5pt}

\section{\ourname Evaluation Framework}
\label{sec:evaluation_framework}
\vspace{-5pt}
We designed our evaluation framework for \ourname and the baseline models with the following objectives: (1) \textbf{Reproducibility}: Our evaluations should be fully reproducible to ensure reliable comparisons. (2) \textbf{Comprehensive coverage}: Our evaluations should encompass diverse test domains (e.g., medicine, physics, and chemistry) and question types (e.g., multiple-choice questions and computational problems). (3) \textbf{Comparison fairness}: Our evaluation setup, including templates and prompting strategies, should provide equitable treatment across different models. (4) \textbf{Accurate answer extraction}: Our evaluation should reliably extract answers from model responses, as the answer extraction methodology significantly impacts final accuracy metrics.

Accordingly, our framework consists of four key components: an open evaluation toolkit for reproducible evaluations (\S~\ref{sec:openscience_eval_system}), a comprehensive suite for evaluating the scientific reasoning abilities of LLMs (\S~\ref{sec:eval_suite}), a series of answer extraction strategies (\S~\ref{sec:answer_extraction}), and a set of recommended evaluation settings based on our experiments with various models (Table~\ref{tab:eval_setting}).

\subsection{Language Model Open Science Evaluation}
\label{sec:openscience_eval_system}
To promote standardized and reproducible evaluations, we are open-sourcing the codebase used to conduct all evaluations in this work~\footnote{\href{https://github.com/GAIR-NLP/lm-open-science-evaluation}{https://github.com/GAIR-NLP/lm-open-science-evaluation}}. Our open science evaluation system offers the following features:

\begin{itemize}[leftmargin=10pt]
    \item Support for both conversation models and base models;
    \item Easy integration of new benchmarks and configurations (e.g., prompting and few-shot settings);
    \item Scalable evaluation of multiple models, benchmarks, and tasks in a single run with multi-node and multi-GPU parallelization;
    \item Comprehensive instance-level output data enabling fine-grained analysis of model predictions.
\end{itemize}

\subsection{\ourname Evaluation Suite}
\label{sec:eval_suite}
To comprehensively evaluate scientific abilities, our evaluation framework encompasses both general science knowledge and specialized subject areas across multiple question formats. Below, we introduce our category and the included benchmarks.

\begin{itemize}[leftmargin=10pt]
    \item \textbf{General Scientific Reasoning:} MMLU~\citep{hendrycks2020mmlu}, GPQA-Diamond~\citep{rein2024gpqa}, MMLU-Pro~\citep{wang2024mmlupro}, SuperGPQA~\citep{du2025supergpqa}, SciBench~\citep{wang2023scibench}, and OlympicArena~\citep{huang2024olympicarena}.
    \item \textbf{Specific Scientific Reasoning:} ChemBench~\citep{mirza2024chembench}, CS-Bench~\citep{song2024csbench}, MedQA~\citep{jin2020medqa}, MedMCQA~\citep{pal2022medmcqa}, PubMedQA~\citep{jin2019pubmedqa}, and PIQA~\citep{bisk2020piqa}.
    \item \textbf{Mathematic Reasoning:} GSM8K~\citep{cobbe2021gsm8k}, MATH~\citep{hendrycks2021math}, and MATH500~\citep{lightman2023letverifystepbystep}.
\end{itemize}
















\begin{table}[t]
    \caption{The \ourname evaluation settings. \textbf{CoT} denotes evaluations conducted with chain-of-thought prompting. \textbf{Unit} indicates that the answer requires unit assignment. \textbf{EM (unit)} represents exact match accuracy for both the numerical answer and its corresponding unit.}
    \centering
    \resizebox{\textwidth}{!}{ 
    \begin{tabular}{l|lcccc}
    \toprule
    \textbf{Category} & \textbf{Benchmark} & \textbf{Question Type} & \textbf{CoT} & \textbf{Unit} & \textbf{Metric} \\
    \midrule
    \multirow{6}{*}{General Reasoning} & MMLU & Multi-Choice & \ding{51} & \ding{55} & EM \\
    & GPQA-Diamond &  Multi-Choice & \ding{51} & \ding{55} & EM \\
    & MMLU-Pro &  Multi-Choice & \ding{51} & \ding{55} & EM \\
    & SuperGPQA &  Multi-Choice & \ding{51} & \ding{55} & EM \\
    & SciBench & Computational Problems & \ding{51} & \ding{51} & EM (unit) \\
    & OlympicArena & Computational Problems & \ding{51} & \ding{51} & EM (unit) \\
    \midrule
    Chemistry & ChemBench & Multi-Choice  \& Problem-Solving  & \ding{51} & \ding{55} & EM \\
    \midrule
    Computer Science & CS-Bench & Multi-Choice \& True/False & \ding{51} & \ding{55} & EM \\
    \midrule
    \multirow{3}{*}{Medicine} & MedQA & Multi-Choice & \ding{51} & \ding{55} & EM \\
    & MedMCQA & Multi-Choice & \ding{51} & \ding{55} & EM \\
    & PubMedQA & Multi-Choice & \ding{51} & \ding{55} & EM \\
    \midrule
     Physics & PIQA & Multi-Choice  & \ding{51} & \ding{55} & EM \\
    \midrule
    \multirow{3}{*}{Math} & GSM8K & Computational Problems & \ding{51} & \ding{55} & EM \\
    & MATH & Computational Problems & \ding{51} & \ding{55} & EM \\
    & MATH500 & Computational Problems & \ding{51} & \ding{55} & EM \\
    \bottomrule
    \end{tabular}
    }
    \label{tab:eval_setting}
\end{table}

\subsection{Answer Extraction Strategy}
\label{sec:answer_extraction}
Answer extraction is critically important for evaluation, as extraction accuracy can substantially impact overall results. Many scientific evaluations simply extract content within \texttt{\textbackslash boxed\{\}}, often omitting responses that lack this formatting and incorrectly attributing such formatting errors to reduced overall accuracy. To enhance extraction precision, we develop a comprehensive set of rule-based methods tailored to extract answers across diverse question types. Our answer extraction method operates through a two-stage process: (1) identifying answer indicator phrases that signal the presence of a final answer, and (2) extracting the answer content from various formatting patterns. For answer indicators, we recognize patterns such as \texttt{The final answer to this question is <ANSWER>} and \texttt{The correct answer is <ANSWER>}. For answer formats, we handle multiple mathematical and textual formatting styles including \texttt{\textbackslash boxed\{\}}, \texttt{\textbackslash mathrm\{\}}, and \texttt{\textbackslash mathbf\{\}}. The complete set of extraction rules is provided in Table \ref{tab:answer_extraction}. Moreover, for multiple-choice questions, we search the option content and match the corresponding option label if direct extraction of the option label fails.

\section{Supervised Finetuning}
\label{sec:supervised_finetuning}
We conduct supervised fine-tuning to verify the effectiveness of \ourdataset and \megascience, and demonstrate the impact of each component in our data curation pipeline through comprehensive ablation studies.

\subsection{Setup}

\paragraph{Baselines}
We compare our datasets to other scientific reasoning datasets, including:
\begin{itemize}[leftmargin=10pt]
    \item \textbf{SCP-116K}~\citep{lu2025scp} is a science problem and solution dataset consisting of 274K instances, including questions scraped from Web and long-thought solutions generated by DeepSeek-R1.
    \item \textbf{NaturalReasoning}~\citep{yuan2025naturalreasoning} is a general reasoning dataset consisting of 1.1M instances synthesized by Llama3.3-70B-instruct and grounded in web sources, covering math, STEM, economics, social sciences, and other subjects.
    \item \textbf{Nemotron-Science}~\citep{bercovich2025nemotron} is a diverse dataset comprising 708K instances of open-ended and multiple-choice questions (MCQs). The dataset combines questions extracted from StackOverflow with synthetically generated MCQs. Solutions are generated using DeepSeek-R1 and subsequently filtered through rejection sampling to select correct answers.
\end{itemize}

Since these baselines rely on n-gram overlap methods for benchmark decontamination, which can be easily circumvented by minor textual variations and thus fail to ensure the integrity of benchmark evaluations, we apply LLM-based benchmark decontamination (detailed in \S\ref{sec:benchmark_decontamination}) to these baseline datasets to ensure fair comparison. Our LLM-based decontamination approach identified 19K instances of benchmark leakage in SCP-116K, 66K instances in NaturalReasoning, and 164K instances in Nemotron-Science, demonstrating the limitations of n-gram-based benchmark decontamination methods.

\paragraph{Evaluation}
We employ our Language Model Open Science Evaluation to evaluate scientific reasoning abilities; the details of the evaluation framework are described in \S\ref{sec:evaluation_framework}\footnote{MMLU is excluded from our evaluation due to its limited difficulty, which renders it inadequate for evaluating advanced reasoning abilities.}.

\paragraph{Training Details} We use LLaMA-Factory~\citep{zheng2024llamafactory} to fine-tune base models including Qwen2.5, Qwen3, and Llama3 series on our datasets and baselines. The hyperparameters are shown in Table~\ref{tab:superparameters_sft}. Unless otherwise specified, all experiments are conducted on Qwen2.5-7B.


\subsection{Main Experiments}
\begin{table}[t]
    \caption{The main results for scientific reasoning. All models are trained on Qwen2.5-7B. \textbf{DC} indicates LLM-based question decontamination. \textbf{Bold} indicates the best and \underline{underline} indicate the second-best results.}
    \centering
    \newcolumntype{C}{>{\columncolor{MegaScience!20}}c}
    \resizebox{\textwidth}{!}{ 
    \begin{tabular}{ll|c|ccc|CC}
    \toprule
    \textbf{Subject} & \textbf{Benchmark} & \makecell{\textbf{Qwen2.5-7B}\\\textbf{Instruct}} & \makecell{\textbf{SCP-116K}\\\textbf{-DC}} & \makecell{\textbf{Natural}\\\textbf{Reasoning}\\\textbf{-DC}} & \makecell{\textbf{Nemotron}\\\textbf{Science}\\\textbf{-DC}} & \makecell{\textbf{\textsc{Textbook}}\\\textbf{\textsc{Reasoning}}} & \makecell{\textbf{\textsc{Mega}}\\\textbf{\textsc{Science}}} \\
    \midrule
    \multirow{5}{*}{General} & MMLU-Pro & 56.23 & 57.75 & 52.80 & \textbf{62.87}  & 55.48 & \underline{59.16} \\
    & GPQA-D & 31.31 & 29.80 & 31.31 & 29.29  & \underline{34.34} & \textbf{36.36} \\
    & SuperGPQA & 28.78 & 29.81 & 25.84 & \underline{31.06}  & 29.64 & \textbf{31.52} \\
    & SciBench & 42.97 & 28.60 & 40.78  & 23.44   & \underline{44.06} & \textbf{48.75} \\
    & OlympicArena & \underline{36.42} & 23.33 & 33.61 & 29.14  & 34.37 & \textbf{40.23} \\
    \midrule
    Chemistry & ChemBench & 51.90 & 45.55 & \underline{52.58} & 44.37  & 50.97 & \textbf{53.48} \\
    \midrule
    CS & CS-Bench & \underline{69.51} & 66.71 & 68.16  & \textbf{72.21}  & 68.79 & 68.73 \\
    \midrule
    \multirow{3}{*}{Medicine} & MedQA & 54.28 & 50.27 & 56.56 & \textbf{65.28}  & 55.85 & \underline{60.97} \\
    & MedMCQA & 55.87 & 52.47 & 54.86 & \underline{58.47}  & 56.25 & \underline{57.35} \\
    & PubMedQA & 73.60 & 63.40  & \underline{74.20} & \textbf{76.80}  & 74.00 & 73.00 \\
    \midrule
    Physics & PIQA & \underline{86.67} & 75.30 & 86.40 & \textbf{88.25}  & 85.04 & 85.80\\
    \midrule
    \multirow{3}{*}{Math} & GSM8K & \textbf{91.96} & 86.43 & \underline{91.58} & 80.82  & 89.76 & 89.84 \\
    & MATH & \underline{74.90} & 74.10 & 68.90 & 66.96  & 71.44 & \textbf{76.58} \\
    & MATH500 & \underline{68.80} & 68.00 & 66.60 & 57.20  & 66.60 & \textbf{72.40} \\
    \midrule
    \multicolumn{2}{c|}{Average} & \underline{58.80} & 53.68 & 57.44 & 56.15  & 58.33 & \textbf{61.01} \\
    \bottomrule
    \end{tabular}
    }
    \label{tab:main_results_general_reasoning}
\end{table}

\paragraph{\ourdataset demonstrates superior performance across open-source scientific datasets} Our \ourdataset outperforms other open-source datasets across most benchmarks, particularly excelling in computational reasoning tasks. While Nemotron-Science achieves higher performance on multiple-choice benchmarks such as MMLU-Pro and medicine tasks, this advantage stems from its training data consisting entirely of multiple-choice questions, which creates a distribution bias toward such formats. Conversely, Nemotron-Science shows notable deficiencies in computational tasks. \ourdataset achieves substantial improvements over Nemotron-Science, outperforming it by 20.62\% on SciBench and 5.23\% on OlympicArena, while maintaining competitive results on multiple-choice evaluations with only minor performance gaps.

\paragraph{\megascience achieves state-of-the-art performance} Our \megascience demonstrates superior performance by achieving the best results on 7 out of 14 benchmarks and securing second-best performance on 3 additional benchmarks. The method shows substantial improvements over the baseline Qwen2.5-7B-Instruct, with an overall average improvement of 2.21\%. Notably, \megascience excels across diverse scientific domains, achieving the highest performance on challenging computational tasks such as SciBench (48.75\%) and OlympicArena (40.23\%), while also demonstrating strong performance on specific domain benchmarks.

\subsection{Pushing the Frontier in Science Domain with \megascience}
We demonstrate the broader effectiveness of \megascience by training it on Qwen2.5~\citep{yang2025qwen2.5}, Qwen3~\citep{yang2025qwen3}, and Llama3.1~\citep{grattafiori2024llama3} series base models with the same hyperparameters specified in Table~\ref{tab:superparameters_sft}. Our experimental results reveal three key findings that highlight the potential of \megascience for advancing scientific domain capabilities.

\begin{itemize}[leftmargin=10pt]

\item \textbf{Breaking performance barriers in science domain} Training with \megascience improves performance across different model families and scales. As shown in Table~\ref{tab:push_frontier}, Qwen2.5-7B, all Qwen3 series models, and Llama3.1-8B trained on \megascience substantially outperform their corresponding official instruction-tuned counterparts in average performance. This improvement across diverse base models demonstrates that \megascience can effectively push the frontier in the science domain.

\item \textbf{Scaling benefits for larger and stronger models} We observe that \megascience exhibits greater effectiveness for larger and stronger models, suggesting a scaling benefit for scientific instruction tuning. Within the Qwen2.5 series, we find an interesting non-monotonic pattern: while Qwen2.5-1.5B-Instruct outperforms Qwen2.5-1.5B-MegaScience by 2.99\%, this gap narrows significantly to only 0.15\% for the 3B model, and then reverses dramatically with Qwen2.5-7B-MegaScience achieving a 2.21\% improvement over its instruction-tuned baseline. Furthermore, when comparing across model generations, the superior Qwen3 series shows that MegaScience variants outperform official instruct models across all model sizes, with performance gaps that increase proportionally with model scale.

\item \textbf{Mathematical reasoning requires sufficient model capacity} We identify that mathematical capabilities present a particular challenge that requires sufficient model capacity to benefit from our dataset. Our models only surpass official instruction-tuned models in mathematical reasoning when applied to stronger base models such as Qwen2.5-7B and Qwen3-8B. We hypothesize that this selective improvement stems from the advanced difficulty level of mathematical problems in our dataset, many of which involve undergraduate-level or higher specialized mathematical concepts. Such complex mathematical reasoning appears to require models to reach a certain capability threshold before they can effectively learn from and benefit from this challenging reasoning data.

\end{itemize}
\begin{table}[t]
\caption{Comparison between models trained on \megascience and official instruction-tuned models. \textbf{Bold} indicates the best. For fair comparison, Qwen3 adopts non-thinking mode due to our short CoT. The detailed results are shown in Table~\ref{tab:detailed_push_results_qwen2.5} and~\ref{tab:detailed_push_results_qwen3}.} 
\centering
\small
\resizebox{0.8\textwidth}{!}{
\begin{tabular}{l|ccc|c}
    \toprule
    \textbf{Model} & \textbf{General Avg.} & \textbf{Specific Avg.} & \textbf{Math Avg.} & \textbf{All Avg.} \\
    \midrule

    \multicolumn{5}{c}{\textbf{Llama3.1}} \\
    \midrule
    
    Llama3.1-8B-Instruct & 24.44 & \textbf{64.79} & \textbf{61.49} & 49.67 \\
    Llama3.1-8B-\megascience & \textbf{33.99} & 64.17 & 53.33 & \textbf{51.07} \\
    \midrule
    \multicolumn{5}{c}{\textbf{Qwen2.5}} \\
    \midrule
    Qwen2.5-1.5B-Instruct & \textbf{23.42} & \textbf{53.83} & \textbf{59.50} & \textbf{44.18} \\
    Qwen2.5-1.5B-\megascience & 20.77 &	50.67 &	56.23 &	41.19 \\
    \midrule

    Qwen2.5-3B-Instruct & \textbf{32.31} & 59.38 & 67.72 & \textbf{51.50} \\
    Qwen2.5-3B-\megascience & 30.96 & \textbf{59.80} & \textbf{68.40} & 51.35 \\
    
    \midrule
    Qwen2.5-7B-Instruct & 39.14 & 65.31 & 78.55 & 58.80 \\
    Qwen2.5-7B-\megascience & \textbf{43.20} & \textbf{66.55} & \textbf{79.61} & \textbf{61.01} \\
    
    \midrule
    \multicolumn{5}{c}{\textbf{Qwen3}} \\
    \midrule
    
    Qwen3-1.7B-Instruct & \textbf{32.46} & 52.14 & \textbf{73.82} & 49.76 \\
    Qwen3-1.7B-\megascience & 31.66 & \textbf{57.53} & 68.84 & \textbf{50.71} \\
    
    \midrule
    Qwen3-4B-Instruct & 44.91 & 65.78 & \textbf{84.08} & 62.25 \\
    Qwen3-4B-\megascience & \textbf{45.80} & \textbf{66.83} & 82.34 & \textbf{62.64} \\
    \midrule
    
    Qwen3-8B-Instruct & 50.45 & 69.53 & 84.02 &	65.82 \\
    Qwen3-8B-\megascience & \textbf{52.60} & \textbf{71.43} & \textbf{86.19} & \textbf{67.87}\\

    \midrule
    Qwen3-14B-Instruct & 53.59 & 72.19 & 86.87 & 68.70 \\
    Qwen3-14B-\megascience & \textbf{58.07} & \textbf{74.21} & \textbf{88.54} & \textbf{71.52}\\
     \midrule
     Qwen3-30B-A3B-Instruct & 55.66 & 74.61 & 87.55 & 70.62 \\
    Qwen3-30B-A3B-\megascience & \textbf{61.12} & \textbf{76.75} & \textbf{89.33} & \textbf{73.86} \\
    
    \bottomrule
    \end{tabular}}
    \label{tab:push_frontier}
\end{table}

\subsection{Ablation Study}

\paragraph{Impact of Core Components}
To understand the contribution of core components in the pipeline of \ourdataset, we conduct an ablation study by systematically removing individual components. The results are presented in Table~\ref{tab:ablation_each_component}.
The refinement component is crucial for overall performance. Removing it (w/o Refinement) causes a dramatic drop from 58.33\% to 13.15\% overall average, highlighting its critical importance in generating high-quality reasoning steps. The supplementary CoT component also contributes meaningfully, with its removal (w/o Supplementary CoT) decreasing overall performance to 57.33\%. This indicates that providing complete solutions in the answers is essential for enhancing the model's reasoning capabilities, as the detailed step-by-step guidance helps the model learn more effective reasoning patterns.
The decontamination process demonstrates its effectiveness by the expected performance improvements when removed (w/o Decontamination): overall average increases to 58.57\%, confirming that our LLM-based decontamination successfully identifies and removes potentially contaminated examples for more rigorous evaluation.

\paragraph{Impact of Different Models for Refinement}
The results in Table~\ref{tab:ablation_different_refine_model} demonstrate the impact of using different models for QA refinement in \ourdataset. DeepSeek-V3 consistently outperforms Llama3.3-70B-Instruct across all evaluation categories, indicating that employing more capable models for data refinement leads to improved downstream performance, suggesting that the quality of the refinement process is directly correlated with the sophistication of the underlying refinement model.
\begin{table}[t]
\centering
\begin{minipage}{0.6\textwidth}
\centering
\caption{The impact of each component. \textbf{Bold} indicates the best results.}
\label{tab:ablation_each_component}
\small
\resizebox{\textwidth}{!}{
\begin{tabular}{l|ccc|c}
    \toprule
    \textbf{Dataset} & \textbf{General Avg.} & \textbf{Specific Avg.} & \textbf{Math Avg.} & \textbf{All Avg.} \\
    \midrule
    \rowcolor{MegaScience!20}
    \textsc{TextbookReasoning} & 39.58 & \textbf{65.15} & 75.93 & 58.33\\
    \midrule
    w/o Decontamination & \textbf{39.87} & 65.12 & \textbf{76.65} & \textbf{58.57}  \\
    w/o Supplementary CoT & 37.63 & 64.54 & 75.73 & 57.33 \\
    w/o Refinement & \textcolor{white}{0}4.32 & 20.37 & 13.42 & 13.15\\
    \bottomrule
\end{tabular}}
\end{minipage}
\hfill
\begin{minipage}{0.38\textwidth}
\centering
\caption{The impact of different models of refinement. \textbf{Bold} indicates the best results.}
\label{tab:ablation_different_refine_model}
\small
\resizebox{\textwidth}{!}{
\begin{tabular}{l|cc}
    \toprule
    \textbf{Results} & \makecell{\textbf{Llama3.3-70B}\\\textbf{-Instruct}} & \makecell{\textbf{DeepSeek}\\\textbf{-V3}}  \\
    \midrule
    General Avg. & 34.23 & \textbf{37.63} \\
    Specific Avg. & 63.84  &  \textbf{64.54} \\
    Math Avg. & 74.26 & \textbf{75.73} \\
    All Avg. & 55.50 & \textbf{58.33} \\
    \bottomrule
    \end{tabular}}
\end{minipage}
\end{table}




\vspace{1cm}

\subsection{Analysis}
\paragraph{Impact of Decontamination}
Existing datasets primarily employ n-gram based decontamination methods, which can be easily circumvented by minor variations in phrasing or structure. To address this limitation, we applied LLM-based question decontamination~\citep{toshniwal2024openmathinstruct, he2025deepmath} to all datasets used in our experiments (see \S\ref{sec:benchmark_decontamination} for details). 
\begin{wraptable}{r}{0.6\textwidth}
\caption{The impact of LLM-based question decontamination. \textbf{Bold} indicates the best results.}
\centering
\small
\resizebox{0.6\textwidth}{!}{
\begin{tabular}{l|ccc|c}
    \toprule
    \textbf{Dataset} & \textbf{General Avg.} & \textbf{Specific Avg.} & \textbf{Math Avg.} & \textbf{All Avg.} \\
    \midrule
    SCP-116K & \textbf{35.76} & \textbf{60.29} & \textbf{77.93} & \textbf{55.31} \\
    + Decontamination & 33.86 & 58.95 & 76.18 & 53.68 \\
    \midrule
    NaturalReasoning & 36.60 & \textbf{65.77} & 74.08 & 57.13 \\
    + Decontamination & \textbf{36.87} & 65.46 & \textbf{75.69} & \textbf{57.44} \\
    \midrule
    Nemotron-Science & \textbf{35.79} & \textbf{67.60} & \textbf{69.30} & \textbf{56.60} \\
    + Decontamination & 35.16 & 67.56 & 68.33 & 56.15\\
    \midrule
    \textsc{TextbookReasoning} & \textbf{39.87} & 65.12 & \textbf{76.65} & \textbf{58.57}\\
    + Decontamination & 39.58 & \textbf{65.15} & 75.93 & 58.33 \\
    \bottomrule
    \end{tabular}}
    \label{tab:decontamination_performance}
\end{wraptable}

Table~\ref{tab:decontamination_performance} presents the results of this decontamination process across the four datasets. We observe varying impacts of LLM-based decontamination, with three of the four datasets demonstrating performance degradation after decontamination, confirming the effectiveness of our approach in identifying and removing contaminated samples. SCP-116K exhibits the most substantial performance drop, indicating a relatively high level of data contamination in this dataset. Nemotron-Science also shows modest decreases across benchmarks, suggesting the presence of contaminated samples that artificially inflated the original performance.
In contrast, NaturalReasoning presents an upward trend after decontamination, suggesting that NaturalReasoning has a lower contamination rate.


\begin{wrapfigure}{r}{0.45\textwidth}
    \centering
    \includegraphics[width=0.45\textwidth]{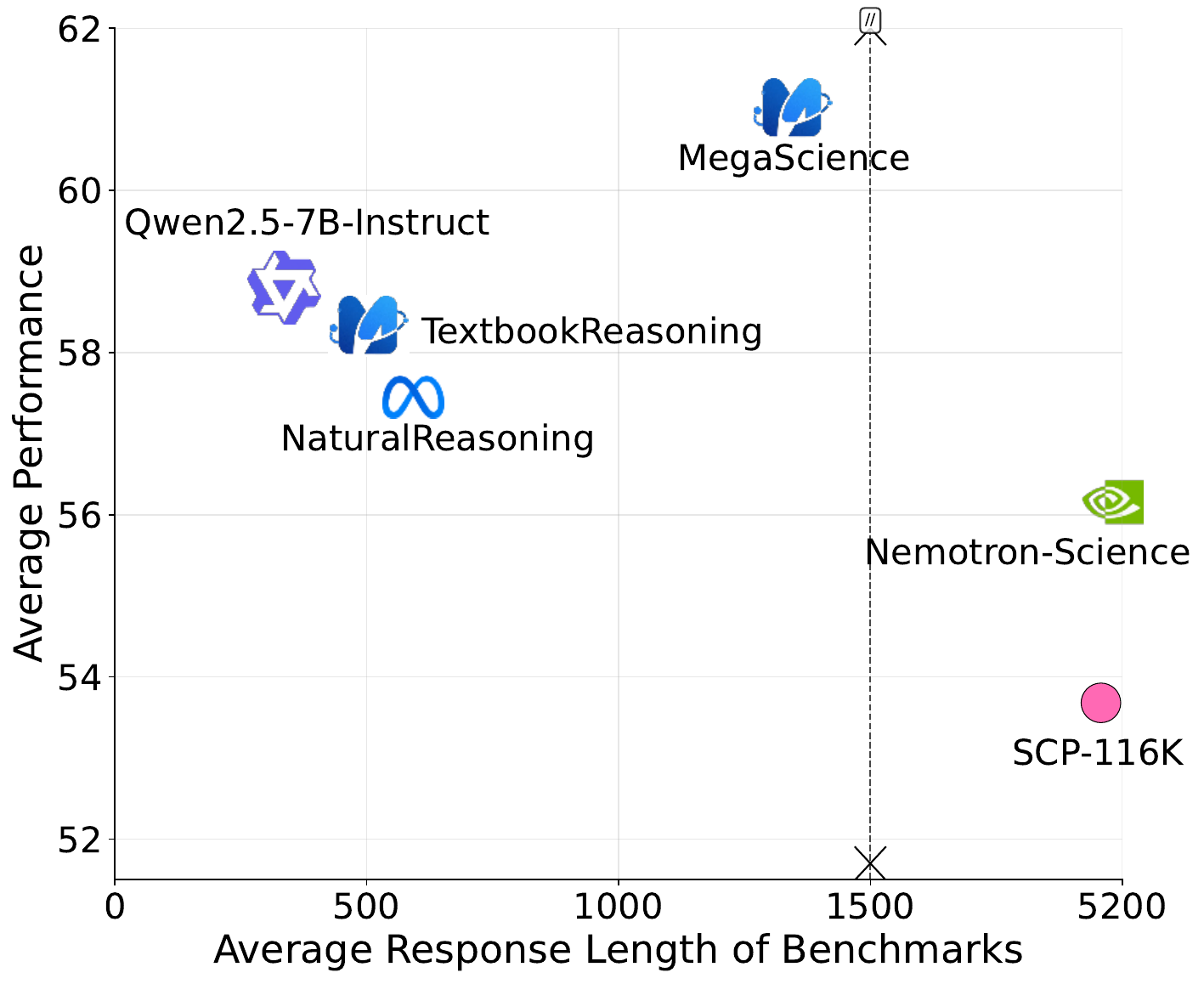}
    \caption{Trade-off between model performance and average response length of all benchmarks. The upper-left region indicates datasets that achieve high performance with better efficiency.}
    \label{fig:inference_response_length_vs_performance}
\end{wrapfigure}
\paragraph{Performance-Efficiency Trade-off Analysis}
A fundamental challenge in reasoning model development lies in balancing performance and efficiency. While recent reasoning models employ long CoT to improve performance, our analysis reveals a \emph{counterintuitive phenomenon} in existing open-source scientific reasoning datasets. 
(1) To investigate the relationship between training efficiency and performance, we compare the average response length of training datasets with the downstream performance of Qwen2.5-7B models trained on them. As illustrated in Figure~\ref{fig:response_length_vs_performance}, we observe a negative correlation: longer training responses often lead to worse performance, which we attribute to poor question quality and difficulty. This explains why naive distillation from models like DeepSeek-R1, despite producing long CoTs, fails to yield satisfactory results—resulting in solutions that are neither performant nor efficient. In contrast, our high-quality dataset \ourdataset achieves the best trade-off, appearing in the upper-left region and demonstrating that carefully curated short CoT can support both strong performance and training efficiency. 
(2) To further examine the inference efficiency–performance trade-off, we analyze the relationship between the overall average response length across all benchmarks and the corresponding average performance during inference. As shown in Figure~\ref{fig:inference_response_length_vs_performance}, models trained on \megascience, despite using shorter training responses, exhibit strong generalization during inference: models trained on short CoT responses of \megascience can elicit long and detailed reasoning. This dynamic adaptation leads to higher average response length during evaluation and, crucially, a substantial boost in performance—highlighting that efficiency at training time does not preclude flexible and effective reasoning at inference time. Furthermore, the average inference response length of Qwen3-8B-\megascience (1080 tokens) is shorter than that of Qwen2.5-7B-\megascience (1345 tokens), suggesting that more advanced models are capable of producing more concise and efficient outputs.


\begin{wraptable}{r}{0.45\textwidth}
\caption{Comparison of difficulty-aware distillation and refinement approaches using DeepSeek-V3 across both datasets. \textbf{Bold} indicates the best.}
\centering
\small
\resizebox{0.45\textwidth}{!}{
\begin{tabular}{l|cc}
    \toprule
    \textbf{Results} & \textbf{Distillation} & \textbf{Refinement}  \\
    \midrule
    General Avg. & 38.84 & \textbf{39.58} \\
    Specific Avg. & \textbf{65.43}  & 65.15 \\
    Math Avg. & \textbf{76.39} & 75.93 \\
    All Avg. & 58.28 & \textbf{58.33} \\
    \bottomrule
    \end{tabular}}
    \label{tab:distillation_vs_refinement}
\end{wraptable}
\paragraph{Comparison Between Difficulty-Aware Distillation and Refinement}
To investigate whether distilling long CoT reasoning specifically for difficult problems yields better performance than refined answers, we applied difficulty selection (see \S\ref{sec:data_selection}) to \ourdataset, identifying 55k problems with average scores below 6 as challenging examples. 
We then employed DeepSeek-V3 to generate step-by-step solutions for these questions and compared them against the original refined answers. As shown in Table \ref{tab:distillation_vs_refinement}, refinement achieves slightly better overall performance than difficulty-aware distillation. This advantage likely stems from refinement having access to reference documents that reduce hallucinations, while distillation, despite generating longer CoT reasoning, relies solely on the model's internal knowledge and is more susceptible to hallucinations. Notably, distillation demonstrates a significant improvement in mathematical reasoning tasks, suggesting that long CoT is particularly beneficial for mathematics.

\begin{wrapfigure}{r}{0.4\textwidth}
    \centering
    \includegraphics[width=0.4\textwidth]{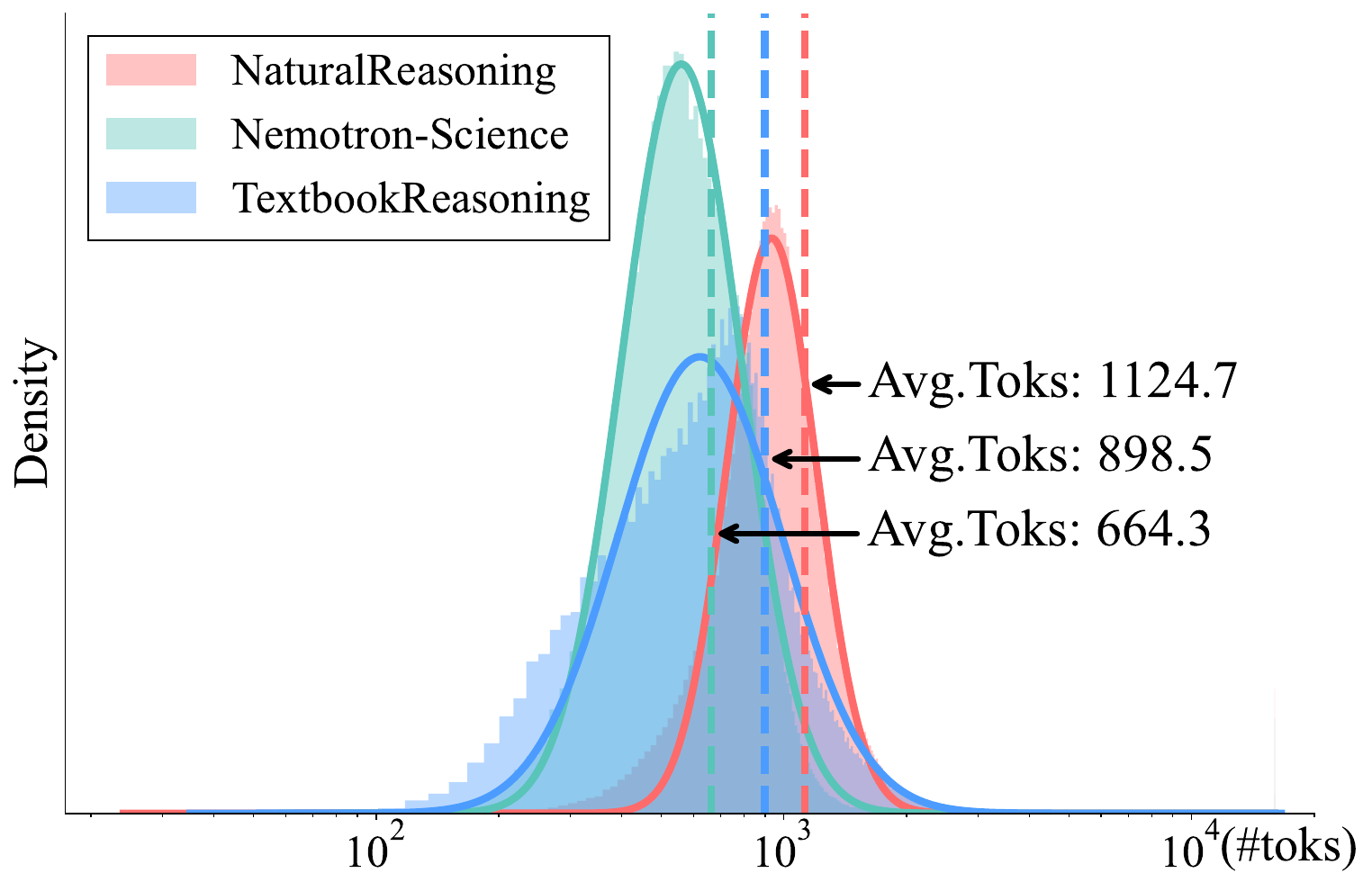}
    \caption{Response token length distributions of Qwen2.5-72B-Instruct across three datasets.}
    \label{fig:dataset_length_distribution}
\end{wrapfigure}
\paragraph{Question Difficulty Analysis}
To estimate question difficulty, we follow \cite{yuan2025naturalreasoning} to leverage a strong LLM (Qwen2.5-72B-Instruct) to generate responses and use response length as a proxy, as longer CoT typically correspond to more complex questions. 
As shown in Figure~\ref{fig:dataset_length_distribution}, while NaturalReasoning exhibits the longest average response length (1124.7 tokens), \ourdataset demonstrates a broader and more diverse difficulty distribution despite having a shorter average length (898.5 tokens). This is evidenced by the wider, flatter probability density curve of \ourdataset, indicating higher variance in response lengths and thus greater diversity in question complexity. In contrast, both NaturalReasoning and Nemotron-Science show more concentrated distributions around their respective means, suggesting more homogeneous difficulty levels within each dataset.

\section{Discussion}

\noindent\textbf{On the Relationship Between Optimal Data Mixture and Model Capability} \quad Our findings reveal that identifying a universally optimal post-training data mixture remains challenging across all base models. Models exhibit significant variations in capacity—whether across different architectures, parameter scales, or generational updates (e.g., Qwen2.5 vs. Qwen3). In this context, such divergence manifests as fundamentally distinct baselines in domain-specific knowledge (e.g., science). Consequently, less capable models—such as Llama series or smaller-scale Qwen2.5 instances—exhibit significant learning struggles when processing complex reasoning datasets like \megascience without supplemental foundational data or lower-difficulty ``warmup'' training. These struggles manifest concretely in suboptimal responses during inference, characterized by abbreviated response length and elevated repetition rates.


\noindent\textbf{The Proxy Model Pitfall in Data Development} \quad When iterating on data quality or studying mixture strategies, reliance on a proxy model for validation is indispensable—yet perilous. In this work, our use of Qwen2.5-7B as a proxy tightly couples experimental outcomes and optimized data mixtures to this specific model’s capabilities. While \megascience data yields significant gains for Qwen2.5-7B, models with lower capacity struggle to replicate these results, necessitating demystification and accessibility adaptations of the data. This underscores a critical caveat: \emph{Proxy model selection inherently biases data development, urging deliberate consideration of capability alignment and broader generalizability in future research.}


\section{Related Works}
The scientific capabilities of LLMs have emerged as a focal point in recent years. With advancements in test-time scaling~\citep{xia2025generativeaiactii}, research focus has shifted from knowledge-based abilities to reasoning capabilities. Current approaches for developing scientific reasoning datasets primarily fall into two categories.

The first approach involves scraping questions from the Web~\citep{lu2025scp, yuan2025naturalreasoning, ma2025general, guha2025openthoughts, li2025naturalthoughts}, where answers can be directly extracted from documents, generated by LLMs provided with relevant documents, or produced through reasoning models such as DeepSeek-R1. The second approach utilizes LLMs to synthesize questions and solutions from seed data~\citep{bercovich2025nemotron}.
However, these existing methods face several critical limitations. First, they struggle to generate high-quality reference answers due to LLMs' hallucination issues. Second, direct distillation from reasoning models leads to overthinking and inefficiency in both training and inference processes. Third, these approaches typically employ only n-gram decontamination, which can be easily circumvented by minor variations in phrasing or structure. Finally, most existing work focuses exclusively on multi-choice benchmarks (e.g., MMLU~\citep{hendrycks2020mmlu}, GPQA~\citep{rein2024gpqa}), which fail to adequately reflect true reasoning abilities such as computational skills, while simultaneously contributing to an overrepresentation of multi-choice questions in training datasets.

To address these limitations, our work introduces several key innovations. First, we adopt textbooks as our primary data source, which provides more reliable content and enables the generation of higher-quality reference answers compared to web-scraped data. Second, we adopt data selection and short CoT annotation by DeepSeek-V3 to achieve superior performance compared to direct distillation from DeepSeek-R1, thereby avoiding the overthinking and inefficiency problems associated with indiscriminate distillation. Third, we implement LLM-based benchmark decontamination across both our datasets and all related datasets, which effectively identifies and excludes data that exhibit semantic similarity to benchmark questions beyond simple n-gram matching. Finally, we design and open-source the Language Model Open Science Evaluation to accelerate progress in scientific reasoning research. This comprehensive evaluation framework encompasses 15 mainstream scientific benchmarks across diverse question types, including multi-choice, computational, true/false, and open-ended problem-solving tasks, thereby providing a more accurate reflection of comprehensive reasoning abilities.

\section{Conclusion and Future Work}
We first introduce \ourdataset, a comprehensive open-source university-level scientific post-training dataset with truthful reference answers, comprising 650k challenging questions and detailed step-by-step solutions from authoritative textbooks. We then present \megascience, the largest collection of high-quality open-source datasets consisting of 1.25 million instances. Through systematic experiments across different data selection methods, we identify optimal curation strategies for each public dataset, providing empirically-grounded guidelines for efficient assembly of high-quality, domain-specific datasets. Supervised finetuning on Qwen-2.5, Qwen-3 and Llama3 series models demonstrates our datasets' effectiveness in pushing the frontier of scientific reasoning, with the resulting models significantly outperforming their official instruct counterparts. We hope that the \textsc{MegaScience} dataset, alongside our released pipeline, evaluation system, and models, will serve as valuable resources and foster further advances in scientific reasoning.

This project opens up several promising directions for future investigation:
\begin{enumerate}[leftmargin=15pt, label=(\arabic*)]
\item While our current work focuses on supervised finetuning, we have not yet explored reinforcement learning for scientific reasoning. Notably, \ourdataset provides reliable reference answers that could serve as high-quality supervision signals for generating reliable rewards in RL frameworks. This foundation presents an excellent opportunity to investigate whether reinforcement learning can further enhance the reasoning capabilities established through our supervised training.

\item Our approach leverages short CoT reasoning during supervised finetuning. A promising direction for future work is to apply RL on top of these SFT models to acquire long CoT reasoning capabilities, thereby examining whether our method can serve as a complementary or even more efficient alternative to conventional mid-training stages~\citep{wang2025octothinker}. If successful, the results would indicate that supervised finetuning on \megascience not only complements mid-training but also offers a more efficient foundation for scaling RL-based approaches toward long CoT reasoning.

\item Due to computing resource constraints, we have not investigated whether compressing long CoT reasoning into more concise formats could achieve better performance at comparable response lengths of \megascience.
\end{enumerate}

\section*{Acknowledgments}
We would like to express our gratitude to Dian Yang for his invaluable support with DeepSeek-v3 inference. We also thank Yang Xiao for his assistance in collecting textbooks during the early stages of our project prototype. We are grateful to Fan Zhou and Xuefeng Li for their helpful discussions throughout this work. Additionally, we acknowledge Lvmanshan Ye for her valuable suggestions regarding color schemes.

\bibliography{iclr2025_conference}
\bibliographystyle{iclr2025_conference}

\clearpage

\appendix
\section{Prompts}
\subsection{Prompts for Q-A Pairs Extraction}
\label{appx_prompts_extraction}
The prompts used for Q-A pair extraction across seven domains (biology, chemistry, computer science, economics, mathematics, medicine, and physics) are presented in Figure~\ref{prompt:high_standard_extracting_biology_qa}--\ref{prompt:low_standard_extracting_physics_qa}.


\subsection{Prompts for Q-A Pairs Refinement}
The prompt used for Q-A pair refinement is shown in Figure~\ref{prompt:qa_refinement}, the prompt for identifying answers that lack chain-of-thought reasoning is shown in Figure~\ref{prompt:identify_no_cot_qa}, and the prompt for filtering defective Q-A pairs is shown in Figure~\ref{prompt:filter_defective_qa}.

\subsection{Prompts for Question Decontamination}
The prompt used for LLM-based question decontamination is shown in Figure~\ref{prompt:llm_judge_for_decontamination}.

\subsection{Prompts for Difficulty Selection}
The prompt used for annotating reference answers is shown in Figure~\ref{prompt:annotate_reference_answer}, and the prompt used for evaluating student answers is shown in Figure~\ref{prompt:evaluating_student_answer_with_reference}.

\section{Answer Extraction Rules and Patterns}
The answer extraction patterns we designed are shown in Table~\ref{tab:answer_extraction}.

\section{Training Details}
The training details is shown in Table~\ref{tab:superparameters_sft}.

\section{Detailed Results}
The detailed results of \megascience are shown in Table~\ref{tab:detailed_push_results_qwen2.5} and~\ref{tab:detailed_push_results_qwen3}.
\begin{table}[t]
    \caption{The detailed results of Llama3.1 and Qwen2.5 models trained on \megascience and official instruction-tuned models. \textbf{Bold} indicates the best.}
    \centering
    \newcolumntype{C}{>{\columncolor{MegaScience!20}}c}
    \resizebox{\textwidth}{!}{ 
    \begin{tabular}{l|cc|cc|cc|cc}
    \toprule
    \textbf{Benchmark} & \makecell{\textbf{Llama3.1}\\\textbf{8B}\\\textbf{instruct}} & \makecell{\textbf{Llama3.1}\\\textbf{8B}\\\textbf{\megascience}} & \makecell{\textbf{Qwen2.5}\\\textbf{1.5B}\\\textbf{instruct}} & \makecell{\textbf{Qwen2.5}\\\textbf{1.5B}\\\textbf{\megascience}}& \makecell{\textbf{Qwen2.5}\\\textbf{3B}\\\textbf{instruct}} & \makecell{\textbf{Qwen2.5}\\\textbf{3B}\\\textbf{\megascience}}& \makecell{\textbf{Qwen2.5}\\\textbf{7B}\\\textbf{instruct}} & \makecell{\textbf{Qwen2.5}\\\textbf{7B}\\\textbf{\megascience}} \\
    \midrule
    MMLU-Pro &  45.15 & \textbf{50.03} & 30.47 & \textbf{34.79} & \textbf{45.20} & 44.91 & 56.23 & \textbf{59.16}\\
    GPQA-D & 24.24 & \textbf{33.33} & \textbf{30.30} & 15.15 & \textbf{32.32} & 24.75 & 31.31 & \textbf{36.36} \\
    SuperGPQA & 19.72 & \textbf{25.56} & \textbf{18.90} & 17.81 & \textbf{23.42} & 22.47 & 28.78 & \textbf{31.52} \\
    SciBench & 10.78 & \textbf{34.06} & 17.81 & \textbf{18.75} & 33.12 & \textbf{36.09} & 42.97 & \textbf{48.75} \\
    OlympicArena & 22.31 & \textbf{26.98} & \textbf{19.62} & 17.36 & \textbf{27.46} & 26.60 & 36.42 & \textbf{40.23} \\

    ChemBench & 49.57 & \textbf{50.39} & \textbf{42.03} & 41.99 & 46.52 & \textbf{47.63} & 51.90 & \textbf{53.48} \\

    CS-Bench & 57.87 & \textbf{59.62} & \textbf{56.91} & 54.61 & \textbf{64.90} & 62.82 & \textbf{69.51} & 68.73 \\

    MedQA & \textbf{67.01} & 60.49 & 37.71 & \textbf{39.36} & \textbf{46.82} & 45.33 & 54.28 & \textbf{60.97} \\
    MedMCQA & \textbf{57.92} & 54.08 & 41.31 & \textbf{43.13} & 48.36 & \textbf{50.51} & 55.87 & \textbf{57.35} \\
    PubMedQA & \textbf{78.80} & 76.80 & \textbf{68.80} & 68.20 & 67.20 & \textbf{71.20} & \textbf{73.60} & 73.00 \\

    PIQA & 77.58 & \textbf{83.62} & \textbf{76.22} & 56.75 & \textbf{82.48} & 81.34 & \textbf{86.67} & 85.8 \\

    GSM8K & \textbf{83.40} & 72.10 & \textbf{73.84} & 72.86 & 80.67 & \textbf{83.02} & \textbf{91.96} & 89.84 \\
    MATH & \textbf{50.48} & 46.90 & \textbf{54.66} & 49.24 & \textbf{65.68} & 62.18 & 74.90 & \textbf{76.58} \\
    MATH500 & \textbf{50.60} & 41.00 & \textbf{50.00} & 46.60 & 56.80 & \textbf{60.00} & 68.80 & \textbf{72.40} \\
    \midrule
    Average & 49.67 & \textbf{51.07} & \textbf{44.18} & 41.19 & \textbf{51.50} & 51.35 & 58.80 & \textbf{61.01}\\
    \bottomrule
    \end{tabular}
    }
    \label{tab:detailed_push_results_qwen2.5}
\end{table}

\begin{table}[t]
    \caption{The detailed results of Qwen3 series models trained on \megascience and official instruction-tuned models. \textbf{Bold} indicates the best. For fair comparison, Qwen3 adopts non-thinking mode due to our short CoT.}
    \centering
    \newcolumntype{C}{>{\columncolor{MegaScience!20}}c}
    \resizebox{\textwidth}{!}{ 
    \begin{tabular}{l|cc|cc|cc|cc|cc}
    \toprule
    \textbf{Benchmark} & \makecell{\textbf{Qwen3}\\\textbf{1.7B}\\\textbf{instruct}} & \makecell{\textbf{Qwen3}\\\textbf{1.7B}\\\textbf{\megascience}} & \makecell{\textbf{Qwen3}\\\textbf{4B}\\\textbf{instruct}} & \makecell{\textbf{Qwen3}\\\textbf{4B}\\\textbf{\megascience}}& \makecell{\textbf{Qwen3}\\\textbf{8B}\\\textbf{instruct}} & \makecell{\textbf{Qwen3}\\\textbf{8B}\\\textbf{\megascience}}& \makecell{\textbf{Qwen3}\\\textbf{14B}\\\textbf{instruct}} & \makecell{\textbf{Qwen3}\\\textbf{14B}\\\textbf{\megascience}}& \makecell{\textbf{Qwen3}\\\textbf{30B-A3B}\\\textbf{instruct}} & \makecell{\textbf{Qwen3}\\\textbf{30B-A3B}\\\textbf{\megascience}} \\
    \midrule
    MMLU-Pro & 40.87 & \textbf{43.94} & 59.42 & \textbf{60.81} & 64.89 & \textbf{66.81} & 68.61 & \textbf{71.60} & 71.78 & \textbf{73.06} \\
    GPQA-D & \textbf{33.33} & 23.23 & \textbf{37.37} & 34.85 & \textbf{47.47} & 46.46 & 49.49 & \textbf{50.51} & 52.02 & \textbf{57.58} \\
    SuperGPQA & \textbf{22.86} & 22.27 & 31.42 & \textbf{33.08} & 35.70 & \textbf{38.84} & 39.87 & \textbf{44.35} & 42.06 & \textbf{46.86} \\
    SciBench & 33.05 & \textbf{41.09} & 51.88 & \textbf{55.00} & 56.41 & \textbf{61.25} & 58.44 & \textbf{68.13} & 59.53 & \textbf{69.22} \\
    OlympicArena & \textbf{32.18} & 27.77 & 44.44 & \textbf{45.25} & 47.79 & \textbf{49.65} & 51.55 & \textbf{55.76} & 52.89 & \textbf{58.86} \\

     ChemBench & 44.33 & \textbf{46.63} & \textbf{54.19} & 54.12 & 54.38 & \textbf{56.78} & 58.07 & \textbf{58.71} & 59.97 & \textbf{61.65} \\

    CS-Bench & 51.52 & \textbf{60.86} & \textbf{70.92} & 70.59 & 74.69 & \textbf{76.43} & 78.18 & \textbf{79.92} & 79.08 & \textbf{81.33} \\

    MedQA & 39.75 & \textbf{43.05} & 57.34 & \textbf{58.84} & 65.99 & \textbf{66.06} & 70.38 & \textbf{71.56} & 76.04 & \textbf{78.16} \\
    MedMCQA & 42.31 & \textbf{47.62} & 54.79 & \textbf{58.28} & 61.18 & \textbf{63.30} & 64.79 & \textbf{66.79} & 67.68 & \textbf{69.27} \\
    PubMedQA & 69.60 & \textbf{71.40} & 73.60 & \textbf{76.80} & 74.20 & \textbf{77.80} & 73.00 & \textbf{78.20} & 74.20 & \textbf{78.40} \\

    PIQA & 65.34 & \textbf{75.63} & \textbf{83.84} & 82.37 & 86.72 & \textbf{88.19} & 88.74 & \textbf{90.10} & 90.70 & \textbf{91.68} \\

    GSM8K & 82.03 & \textbf{82.41} & \textbf{91.74} & 91.58 & 91.89 & \textbf{93.48} & 93.86 & \textbf{94.77} & 94.62 & \textbf{94.69} \\
    MATH & \textbf{73.22} & 63.90 & \textbf{83.50} & 81.44 & 83.98 & \textbf{85.30} & 86.76 & \textbf{88.24} & 87.24 & \textbf{89.90} \\
    MATH500 & \textbf{66.20} & 60.20 & \textbf{77.00} & 74.00 & 76.20 & \textbf{79.80} & 80.00 & \textbf{82.60} & 80.80 & \textbf{83.40} \\
    \midrule
    Average & 49.76 & \textbf{50.71} & 62.25 & \textbf{62.64} & 65.82 & \textbf{67.87} & 68.70 & \textbf{71.52} & 70.62 & \textbf{73.86} \\
    \bottomrule
    \end{tabular}
    }
    \label{tab:detailed_push_results_qwen3}
\end{table}

\begin{figure}[ht]
\small
\centering
\begin{tcolorbox}[
    colback=MegaScience!20!white,    
    colframe=MegaScience  
]
\begin{Verbatim}[breaklines=true,breaksymbol=,]
Below is a biology document extract. Assess whether it contains a biology question-and-answer pair that requires reasoning:

- If the document extract does not contain a biology question-and-answer pair that involves reasoning, return the explicit symbol `[NO QA]`.
- If the document only contains simple factual or conceptual questions without deeper reasoning, return `[NO QA]`.
- If a biology reasoning question-and-answer pair is found, extract it in the following format:
    Question: <question text with complete problem statement and all necessary biological information>
    Answer: <complete solution with all necessary reasoning steps, processes, and explanations included> (only if an answer is provided, otherwise do not generate this line)
- The extracted pair must:
  1. Require logical or scientific reasoning beyond simple recall
  2. Be self-contained and biologically precise
  3. Include all necessary context for independent solving
  4. May involve mechanisms, pathways, evolutionary principles, genetic analysis, experimental design, or systems-level understanding

Do NOT extract simple definitional questions or basic concept recall questions.

#### The extract:
`<DOCUMENT>`

Now process the extract and return the result.
\end{Verbatim}
\end{tcolorbox}
\caption{High-standard prompt for extracting Q-A pairs of biology.}
\label{prompt:high_standard_extracting_biology_qa}
\end{figure}
\begin{figure}[ht]
\small
\centering
\begin{tcolorbox}[
    colback=MegaScience!20!white,    
    colframe=MegaScience  
]
\begin{Verbatim}[breaklines=true,breaksymbol=,]
Below is a book document extract.

# Extraction Task
Extract complete, independently solvable biology questions and answers from the document while following these guidelines:

## For Questions:
- Extract any explicit biology questions with their associated answers
- For implicit biology concepts, mechanisms, processes, or principles presented as statements, convert them to well-formed questions ONLY if they can stand alone
- Ensure each extracted question contains ALL necessary information to be solved independently without requiring additional context
- Include any relevant biological diagrams, pathways, or figures mentioned (describe them if not visible)
- Extract multiple questions separately if they exist
- If no biological content can be meaningfully extracted as a question, return `[NO QA]`

## For Answers:
- Include the answer provided in the extract
- Answers should capture the essential explanation of the biological concept
- If the source material contains a description of a mechanism or pathway, include this in the answer
- For biological processes, the answer should explain the function, steps, or significance as presented in the text

## Format:
Format each question-answer pair as:
Question: [Complete biology question with all context needed to understand]
Answer: [Corresponding answer from the text]

The extract is as follows:
`<DOCUMENT>`

Now process the extract and return the result.
\end{Verbatim}
\end{tcolorbox}
\caption{Low-standard prompt for extracting Q-A pairs of biology.}
\label{prompt:low_standard_extracting_biology_qa}
\end{figure}
\begin{figure}[ht]
\small
\centering
\begin{tcolorbox}[
    colback=MegaScience!20!white,    
    colframe=MegaScience  
]
\begin{Verbatim}[breaklines=true,breaksymbol=,]
Below is a chemistry document extract. Assess whether it contains a chemistry question-and-answer pair requiring significant reasoning:

- If the document extract does not contain a chemistry reasoning question-and-answer pair, return the explicit symbol`[NO QA]`.
- If a chemistry question-and-answer pair requiring reasoning is found, extract it in the following format:
    Question: <question text with complete problem statement and all necessary chemical information>
    Answer: <complete solution with all necessary steps, equations, calculations, and reasoning included> (only if an answer is provided, otherwise do not generate this line)
- The extracted pair must:
  1. Require chemical reasoning or multi-step problem-solving (not simple definition or concept recall)
  2. Be self-contained and chemically precise, allowing independent solving without additional context
  3. Involve topics such as: reaction mechanisms, thermodynamics, equilibrium calculations, acid-base chemistry, electrochemistry, kinetics, spectroscopic analysis, or other areas requiring deductive reasoning
- Do NOT extract simple definitional questions, basic concept recalls, or single-step calculations.

#### The extract:
`<DOCUMENT>`

Now process the extract and return the result.
\end{Verbatim}
\end{tcolorbox}
\caption{High-standard prompt for extracting Q-A pairs of chemistry.}
\label{prompt:high_standard_extracting_chemistry_qa}
\end{figure}
\begin{figure}[ht]
\small
\centering
\begin{tcolorbox}[
    colback=MegaScience!20!white,    
    colframe=MegaScience  
]
\begin{Verbatim}[breaklines=true,breaksymbol=,]
Below is a book document extract.

# Extraction Task
Extract complete, independently solvable chemistry questions and answers from the document while following these guidelines:

## For Questions/Problems:
- Extract any explicit chemistry questions with their answers
- Extract ONLY questions that are completely self-contained and can be solved independently
- For implicit problems (chemical principles, reactions, or concepts presented as statements), convert them to well-formed questions ONLY if they can stand alone
- Ensure each extracted problem contains ALL necessary information to be solved independently
- Include any relevant diagrams, figures, or charts mentioned (describe them if not visible)
- Extract multiple problems separately if they exist
- If no mathematical content can be extracted, return `[NO QA]`

## For Answers:
- Include the complete answer if provided in the extract
- Answers should contain the main solution or explanation
- If a detailed step-by-step solution is available, include it
- For reaction mechanisms, include all steps and intermediates

## Format:
Format each question-answer pair as:
Question: [Complete chemistry question with all context needed to solve]
Answer: [Complete answer]

The extract is as follows:
`<DOCUMENT>`

Now process the extract and return the result.
\end{Verbatim}
\end{tcolorbox}
\caption{Low-standard prompt for extracting Q-A pairs of chemistry.}
\label{prompt:low_standard_extracting_chemistry_qa}
\end{figure}
\begin{figure}[ht]
\small
\centering
\begin{tcolorbox}[
    colback=MegaScience!20!white,    
    colframe=MegaScience  
]
\begin{Verbatim}[breaklines=true,breaksymbol=,]
Below is a document extract. Assess whether it contains a computer science or artificial intelligence question-and-answer pair that requires significant reasoning:

- If the document extract does not contain a computer science or artificial intelligence question-and-answer pair requiring reasoning, return the explicit symbol `[NO QA]`.
- If the extract contains only simple definitional or conceptual questions without reasoning, return the explicit symbol `[NO QA]`.
- If a reasoning-based computer science or artificial intelligence question-and-answer pair is found, extract it in the following format:
    Question: <complete problem statement including all necessary information, constraints, and requirements>
    Answer: <complete solution with all necessary reasoning steps, algorithms, code snippets, or formal proofs> (only if an answer is provided, otherwise do not generate this line)
- The extracted pair must be self-contained and technically precise, allowing independent solving without additional context.
- Prioritize questions that involve algorithm design, computational complexity analysis, system architecture decisions, AI model reasoning, optimization problems, or formal proofs.
- Do not extract simple factual questions about technology history, basic definitions, or conceptual explanations that don't require problem-solving.

#### The extract:
`<DOCUMENT>`

Now process the extract and return the result.
\end{Verbatim}
\end{tcolorbox}
\caption{High-standard prompt for extracting Q-A pairs of computer science and artificial intelligence.}
\label{prompt:high_standard_extracting_cs_qa}
\end{figure}
\begin{figure}[ht]
\small
\centering
\begin{tcolorbox}[
    colback=MegaScience!20!white,    
    colframe=MegaScience  
]
\begin{Verbatim}[breaklines=true,breaksymbol=,]
Below is a book document extract.

# Extraction Task
Extract complete, independently solvable computer science and artificial intelligence questions and answers from the document while following these guidelines:

## For Questions/Problems:
- Extract any explicit computer science or AI questions with their provided answers
- For implicit problems (algorithms, data structures, programming concepts, AI theories, computational theorems, or technical definitions presented as statements), convert them to well-formed questions ONLY if they can stand alone as complete problems
- Ensure each extracted problem contains ALL necessary information to be solved independently without requiring additional context
- Include all context, requirements, constraints, and examples needed to understand the problem
- For computational problems, make sure the question includes all necessary inputs, expected outputs, and constraints
- Extract multiple problems separately if they exist
- If no computer science or AI content can be extracted as complete questions, return `[NO QA]`

## For Answers:
- Include the complete answer as provided in the extract
- Answers should contain the main solution or explanation
- If available, include:
  * Code implementations
  * Time/space complexity analysis
  * Step-by-step explanations
  * Proofs for computational theorems
  * Practical implementation details for AI concepts

## Format:
Format each question-answer pair as:
Question: [Complete computer science/AI question with all context needed to solve]
Answer: [Complete answer]

The extract is as follows:
`<DOCUMENT>`

Now process the extract and return the result.
\end{Verbatim}
\end{tcolorbox}
\caption{Low-standard prompt for extracting Q-A pairs of computer science and artificial intelligence.}
\label{prompt:low_standard_extracting_cs_qa}
\end{figure}
\begin{figure}[ht]
\small
\centering
\begin{tcolorbox}[
    colback=MegaScience!20!white,    
    colframe=MegaScience  
]
\begin{Verbatim}[breaklines=true,breaksymbol=,]
Below is a document extract on economics. Assess whether it contains a challenging economics question-and-answer pair that requires reasoning:

- If the document extract does not contain a challenging economics question-and-answer pair requiring reasoning, return the explicit symbol `[NO QA]`.
- If the document extract contains only simple conceptual definitions or basic knowledge, return `[NO QA]`.
- If a challenging economics question-and-answer pair requiring reasoning is found, extract it in the following format:
    Question: <question text with complete problem statement and all necessary economic information>
    Answer: <complete solution with all necessary reasoning steps, economic analysis, and calculations included> (only if an answer is provided, otherwise do not generate this line)
- The extracted pair must be self-contained and economically precise, allowing independent solving without additional context.

#### The extract:
`<DOCUMENT>`

Now process the extract and return the result.
\end{Verbatim}
\end{tcolorbox}
\caption{High-standard prompt for extracting Q-A pairs of economics.}
\label{prompt:high_standard_extracting_economics_qa}
\end{figure}
\begin{figure}[ht]
\small
\centering
\begin{tcolorbox}[
    colback=MegaScience!20!white,    
    colframe=MegaScience  
]
\begin{Verbatim}[breaklines=true,breaksymbol=,]
Below is a book document extract.

# Extraction Task
Extract complete, independently solvable economics questions and answers from the document while following these guidelines:

## For Questions/Problems:
- Extract any explicit economics questions with their answers
- Extract ONLY questions that are completely self-contained and can be solved independently
- For implicit problems (economic principles, models, theorems, or concepts presented as statements), convert them to well-formed questions ONLY if they can stand alone
- Ensure each extracted problem contains ALL necessary information to be solved independently
- For computational problems (supply/demand analysis, equilibrium pricing, cost-benefit calculations, elasticity, utility maximization, game theory payoffs, etc.), include all required data and parameters
- Include any relevant diagrams, figures, graphs, or tables mentioned (describe them if not visible)
- Extract multiple problems separately if they exist
- If no economics content can be extracted, return `[NO QA]`

## For Answers:
- Include the complete answer if provided in the extract
- Answers should contain the main solution or explanation
- If a detailed step-by-step solution is available, include it
- For model derivations or theoretical proofs, include all steps and reasoning

## Format:
Format each question-answer pair as:
Question: [Complete economics question with all context needed to solve]
Answer: [Complete answer]

The extract is as follows:
`<DOCUMENT>`

Now process the extract and return the result.
\end{Verbatim}
\end{tcolorbox}
\caption{Low-standard prompt for extracting Q-A pairs of economics.}
\label{prompt:low_standard_extracting_economics_qa}
\end{figure}
\begin{figure}[ht]
\small
\centering
\begin{tcolorbox}[
    colback=MegaScience!20!white,    
    colframe=MegaScience  
]
\begin{Verbatim}[breaklines=true,breaksymbol=,]
Below is a math document extract. Assess whether it contains a mathematical question-and-answer pair:

- If the document extract does not contain a mathematical question-and-answer pair, return the explicit symbol`[NO QA]`.
- If a mathematical question-and-answer pair is found, extract it in the following format:
    Question: <question text with complete problem statement and all necessary mathematical information>
    Answer: <complete solution with all necessary steps and calculations included> (only if an answer is provided, otherwise do not generate this line)
- The extracted pair must be self-contained and mathematically precise, allowing independent solving without additional context.

#### The extract:
`<DOCUMENT>`

Now process the extract and return the result.
\end{Verbatim}
\end{tcolorbox}
\caption{High-standard prompt for extracting Q-A pairs of math.}
\label{prompt:high_standard_extracting_math_qa}
\end{figure}
\begin{figure}[ht]
\small
\centering
\begin{tcolorbox}[
    colback=MegaScience!20!white,    
    colframe=MegaScience  
]
\begin{Verbatim}[breaklines=true,breaksymbol=,]
Below is a book document extract.

# Extraction Task
Extract complete, independently solvable mathematical content following these guidelines:

## For Questions/Problems:
- Extract any explicit mathematical questions with their answers
- Convert mathematical theorems, propositions, definitions, or problems without explicit questions into well-formed questions
- Ensure each extracted problem contains ALL necessary information to be solved independently
- Include any relevant diagrams, figures, or charts mentioned (describe them if not visible)
- Extract multiple problems separately if they exist
- If no mathematical content can be extracted, return `[NO QA]`

## For Answers:
- Include the provided solution, proof, or explanation when available
- Brief answers are acceptable if that's all the source provides
- For theorems/propositions, the question should ask to prove the statement

## Format:
Question: <Complete mathematical problem with all context needed to solve>
Answer: <Solution as provided in the extract>

The extract is as follows:
`<DOCUMENT>`

Now process the extract and return the result.
\end{Verbatim}
\end{tcolorbox}
\caption{Low-standard prompt for extracting Q-A pairs of math.}
\label{prompt:low_standard_extracting_math_qa}
\end{figure}
\begin{figure}[ht]
\small
\centering
\begin{tcolorbox}[
    colback=MegaScience!20!white,    
    colframe=MegaScience  
]
\begin{Verbatim}[breaklines=true,breaksymbol=,]
Below is a medical document extract. Assess whether it contains a medical question-and-answer pair that requires clinical reasoning:

- If the document extract does not contain a medical reasoning question-and-answer pair, return the explicit symbol `[NO QA]`.
- If a medical reasoning question-and-answer pair is found, extract it in the following format:
    Question: <question text with complete clinical scenario and all necessary patient information>
    Answer: <complete solution with diagnostic reasoning, differential diagnoses, management plan, and treatment rationale> (only if an answer is provided, otherwise do not generate this line)
- Only extract complex questions requiring clinical reasoning, diagnosis, or treatment planning. Do not extract simple factual or concept-based questions.
- The extracted pair must be self-contained and medically precise, allowing independent assessment without additional context.
- Focus on cases requiring differential diagnosis, interpretation of lab results, management decisions, or therapeutic reasoning.

#### The extract:
`<DOCUMENT>`

Now process the extract and return the result.
\end{Verbatim}
\end{tcolorbox}
\caption{High-standard prompt for extracting Q-A pairs of medicine.}
\label{prompt:high_standard_extracting_medicine_qa}
\end{figure}
\begin{figure}[ht]
\small
\centering
\begin{tcolorbox}[
    colback=MegaScience!20!white,    
    colframe=MegaScience  
]
\begin{Verbatim}[breaklines=true,breaksymbol=,]
Below is a medical document extract.

# Extraction Task
Extract complete, independently solvable medical questions and answers from the document while following these guidelines:

## For Questions:
- Extract any explicit medical questions with their corresponding answers
- For implicit medical cases, conditions, diagnoses, or treatment protocols, convert them into well-formed questions ONLY if they can stand alone
- Ensure each extracted question contains ALL necessary clinical information to be understood and answered independently
- Include any relevant patient data, symptoms, test results, or clinical observations needed to fully understand the case
- Extract multiple questions separately if they exist
- If no medical question content can be extracted, return `[NO QA]`

## For Answers:
- Include the complete answer if provided in the extract
- Focus on capturing the main diagnosis, treatment plan, or clinical explanation
- Answers should be self-contained but don't need to be exhaustive
- Include key points from any detailed explanations or management plans provided

## Format:
Format each question-answer pair as:
Question: [Complete medical question with all context needed to understand the case]
Answer: [Complete answer with diagnosis, treatment, or explanation]

The extract is as follows:
`<DOCUMENT>`

Now process the extract and return the result.
\end{Verbatim}
\end{tcolorbox}
\caption{Low-standard prompt for extracting Q-A pairs of medicine.}
\label{prompt:low_standard_extracting_medicine_qa}
\end{figure}
\begin{figure}[ht]
\small
\centering
\begin{tcolorbox}[
    colback=MegaScience!20!white,    
    colframe=MegaScience  
]
\begin{Verbatim}[breaklines=true,breaksymbol=,]
Below is a physics document extract. Assess whether it contains a physics question-and-answer pair that requires significant reasoning:

- If the document extract does not contain a physics question-and-answer pair requiring substantial reasoning, return the explicit symbol `[NO QA]`.
- If the document extract contains only simple conceptual definitions or basic physics facts without reasoning steps, return `[NO QA]`.
- If a physics question-and-answer pair requiring reasoning is found, extract it in the following format:
    Question: <question text with complete problem statement and all necessary physics information, including any relevant diagrams, equations, or quantities>
    Answer: <complete solution with all necessary reasoning steps, calculations, and physical principles applied> (only if an answer is provided, otherwise do not generate this line)
- The extracted pair must be self-contained and physically precise, allowing independent solving without additional context.

#### The extract:
`<DOCUMENT>`

Now process the extract and return the result.
\end{Verbatim}
\end{tcolorbox}
\caption{High-standard prompt for extracting Q-A pairs of physics.}
\label{prompt:high_standard_extracting_physics_qa}
\end{figure}
\begin{figure}[ht]
\small
\centering
\begin{tcolorbox}[
    colback=MegaScience!20!white,    
    colframe=MegaScience  
]
\begin{Verbatim}[breaklines=true,breaksymbol=,]
Below is a book document extract.

# Extraction Task
Extract complete, independently solvable physics questions and answers from the document while following these guidelines:

## For Questions/Problems:
- Extract any explicit physics questions with their answers
- Extract ONLY questions that are completely self-contained and can be solved independently
- For implicit problems (physics principles, laws, theorems, or concepts presented as statements), convert them to well-formed questions ONLY if they can stand alone
- Ensure each extracted problem contains ALL necessary information to be solved independently
- Include any relevant diagrams, figures, or charts mentioned (describe them if not visible)
- Extract multiple problems separately if they exist
- If no physics content can be extracted, return `[NO QA]`

## For Answers:
- Include the complete answer if provided in the extract
- Answers should contain the key solution or explanation with minimal detail
- For calculation problems, include the relevant formulas, key steps, and final answer with units
- For derivations, include the main steps of the derivation

## Format:
Format each question-answer pair as:
Question: [Complete physics question with all context needed to solve]
Answer: [Complete answer]

The extract is as follows:
`<DOCUMENT>`

Now process the extract and return the result.
\end{Verbatim}
\end{tcolorbox}
\caption{Low-standard prompt for extracting Q-A pairs of physics.}
\label{prompt:low_standard_extracting_physics_qa}
\end{figure}

\begin{figure}[ht]
\small
\centering
\begin{tcolorbox}[
    colback=MegaScience!20!white,    
    colframe=MegaScience  
]
\begin{Verbatim}[breaklines=true,breaksymbol=,]
Below is a question-and-answer pair and a reference document. Your task is to refine the question to make it clear and self-contained, then verify and refine the answer to ensure it's correct and well-explained.

For the question:
- Ensure it contains sufficient information to be understood independently
- Add necessary context from the reference document if the question is unclear
- Maintain the original question's intent

For the answer:
- Verify correctness against the reference document
- If incorrect, provide the correct answer based on the document
- If the answer lacks explanation, add necessary intermediate reasoning process leading to the given answer as a teacher would
- Ensure the added steps are logical, clear, and provide necessary explanation of the solution process
- If the answer already has explanation, reorganize the solution into a clear and well-structured format for better readability and understanding
- For final answers that need exact matching (multiple-choice, calculations, fill-in-the-blank, true/false), use $\\boxed{}$ notation

Requirements:
- The refined question should include all necessary information
- The refined answer should be accurate and well-explained
- Both question and answer should stand alone (no references to documents or original materials)

Output format:
First provide your reasoning for the refinements, then output the final results in this exact format without any notes:

Refined Question: <refined question>
Refined Answer: <refined solution>

I will provide you with the reference document, original question and its answer. Please analyze them carefully before refinement.

The reference document:
`<DOCUMENT>`

The question:
`<PROBLEM>`

The answer:
`<ANSWER>`
\end{Verbatim}
\end{tcolorbox}
\caption{Prompt for refining Q-A pairs.}
\label{prompt:qa_refinement}
\end{figure}
\begin{figure}[ht]
\small
\centering
\begin{tcolorbox}[
    colback=MegaScience!20!white,    
    colframe=MegaScience  
]
\begin{Verbatim}[breaklines=true,breaksymbol=,]
You are an expert evaluator tasked with determining whether an answer contains detailed reasoning processes or explanations of reasons.

**Task**: Given a question and its corresponding answer, analyze whether the answer includes:
- Step-by-step reasoning or logical progression
- Detailed explanations of why something is the case
- Cause-and-effect relationships
- Evidence or justifications for conclusions
- Problem-solving methodology or thought processes

**Instructions**:
1. Carefully read both the question and answer
2. Look for explicit reasoning indicators such as:
   - "Because..." / "Since..." / "Therefore..."
   - Sequential steps (First, Second, Then...)
   - Explanatory phrases ("This is due to...", "The reason is...")
   - Logical connectors and transitions
   - Supporting evidence or examples that explain the reasoning
3. Distinguish between mere factual statements and explanatory reasoning
4. Consider the depth and detail of any reasoning provided

**Output Format**:
Analysis: [Provide your detailed analysis of whether and how the answer demonstrates reasoning or explanation]
Decision: [YES/NO]

**Examples**:

**Example 1:**
Question: Why does ice float on water?
Answer: Ice floats because it is less dense than water. When water freezes, its molecules form a crystalline structure that takes up more space, making ice about 9% less dense than liquid water.

Analysis: The answer provides a clear causal explanation with scientific reasoning. It explains the mechanism (molecular structure change) and quantifies the density difference, showing detailed reasoning about why the phenomenon occurs.
Decision: YES

**Example 2:**
Question: What is the capital of France?
Answer: Paris.

Analysis: This is a simple factual answer without any reasoning process or explanation. It directly states the fact but provides no reasoning about why Paris is the capital or any explanatory context.
Decision: NO

Now analyze the following:
Question: 
`<PROBLEM>`

Answer: 
`<ANSWER>`
\end{Verbatim}
\end{tcolorbox}
\caption{Prompt for identifying answers that lack reasoning processes.}
\label{prompt:identify_no_cot_qa}
\end{figure}
\begin{figure}[ht]
\small
\centering
\begin{tcolorbox}[
    colback=MegaScience!20!white,    
    colframe=MegaScience  
]
\begin{Verbatim}[breaklines=true,breaksymbol=,]
You are tasked with filtering QA (Question-Answer) data to identify problematic entries that should be excluded from a dataset. Please evaluate the provided question and answer pair and determine if it should be filtered out.

## Filtering Criteria

Filter out (mark as NO) any QA pairs that have the following issues:

### 1. Contradictory Answers
The answer contains internal contradictions or conflicting statements.
**Example:**
- Question: What is 2 + 2?
- Answer: First, 2 + 2 = 4. However, using a different method, 2 + 2 = 5. The correct answer is 4.

### 2. External References
The question references external materials that are not provided, such as:
- Specific equations by number (e.g., "equation (8.75)")
- Figures or diagrams (e.g., "as shown in Fig. 4-16")
- External documents or sources not included in the context
**Examples:**
- Question: Solve equation (3.14) using the given parameters.
- Question: Based on Figure 2.1, calculate the area of the triangle.

### 3. Missing or Invalid Answers
The answer does not provide a substantive response to the question, such as:
- Only stating "None of the above" without proper explanation
- Providing no actual answer to the question asked
- Giving completely irrelevant responses
**Example:**
- Question: What is the capital of France?
- Answer: The correct answer is None of the above. This question cannot be answered properly.

## Output Format

After evaluating the question and answer pair, provide your analysis and decision in the following format:

Analysis:
<Provide a brief explanation of your evaluation, noting any issues found or confirming the QA pair is acceptable>

Decision:
<YES/NO>

- YES: Keep this QA pair (it passes the filtering criteria)
- NO: Filter out this QA pair (it has one or more of the issues listed above)

The question:
`<PROBLEM>`

The answer:
`<ANSWER>`
\end{Verbatim}
\end{tcolorbox}
\caption{Prompt for filtering defective Q-A pairs.}
\label{prompt:filter_defective_qa}
\end{figure}

\begin{figure}[ht]
\small
\centering
\begin{tcolorbox}[
    colback=MegaScience!20!white,    
    colframe=MegaScience  
]
\begin{Verbatim}[breaklines=true,breaksymbol=,]
I will now give you two questions: Original question and Candidate question. Please help me determine if the following two questions are the same.

Original question:
`<ORIGINAL_PROBLEM>`

Candidate question:
`<CANDIDATE_PROBLEM>`

Disregard the names and minor changes in word order that appear within.
If their question prompts are very similar and, without considering the solution process, they produce the same answer, we consider them to be the same question.

Output Format:
Analysis: [Provide a detailed analysis evaluating the similarity between these questions]
Decision: [YES/NO]
\end{Verbatim}
\end{tcolorbox}
\caption{LLM prompt for decontamination.}
\label{prompt:llm_judge_for_decontamination}
\end{figure}

\begin{figure}[ht]
\small
\centering
\begin{tcolorbox}[
    colback=MegaScience!20!white,    
    colframe=MegaScience  
]
\begin{Verbatim}[breaklines=true,breaksymbol=,]
## Task Description
You are tasked with extracting the final reference answer from a detailed solution that contains both reasoning steps and the final answer. The reference answer should be concise and represent the definitive conclusion that can be used as a standard solution.

## Input Format
You will receive:
1. A question that was asked
2. A detailed answer that includes reasoning steps and the final answer

## Output Requirements
- Extract ONLY the final reference answer without the reasoning steps
- Ensure the reference answer is complete and can stand alone
- Format the reference answer clearly and concisely
- Do not add any additional explanations or reasoning not present in the original answer
- If multiple possible answers are given, identify the one marked as final or preferred

## Example
### Question:
What is the area of a circle with radius 5 cm?

### Detailed Answer:
To find the area of a circle, I need to use the formula A = πr².
Given information: radius = 5 cm
Substituting values: A = π × 5² = π × 25 = 78.54 cm²
Therefore, the area of the circle with radius 5 cm is 78.54 cm².

### Reference Answer:
78.54 cm²

## Instructions
1. Read the question carefully to understand what is being asked
2. Analyze the detailed answer to identify where the final conclusion is stated
3. Extract only the reference answer without any additional reasoning
4. Format the reference answer clearly so it can be used for checking solutions

## Question:
`<PROBLEM>`

## Detailed Answer:
`<ANSWER>`

Now process and return the result.
\end{Verbatim}
\end{tcolorbox}
\caption{Prompt for annotating reference answer.}
\label{prompt:annotate_reference_answer}
\end{figure}
\begin{figure}[ht]
\small
\centering
\begin{tcolorbox}[
    colback=MegaScience!20!white,    
    colframe=MegaScience  
]
\begin{Verbatim}[breaklines=true,breaksymbol=,]
You are an experienced education evaluator tasked with assessing student responses to academic questions. Your goal is to analyze each student answer in relation to the reference answer and provide both detailed feedback and a numerical score.

Evaluation Process:
1. Carefully read the question to understand the specific requirements and expected knowledge being tested.
2. Compare the student's response to the reference answer, focusing on:
   - Conceptual understanding
   - Accuracy of information
   - Completeness of the answer
   - Use of appropriate terminology
   - Logical reasoning and structure
   - Mathematical correctness (where applicable)

3. Provide a thorough analysis that:
   - Identifies specific strengths in the student's response
   - Points out any errors, misconceptions, or omissions
   - Evaluates how well the answer addresses all parts of the question
   - Considers whether the student demonstrated the required knowledge and skills

4. Assign a score on a scale of 0-10 where:
   - 0: No relevant content or completely incorrect
   - 1-3: Major conceptual errors or significant omissions
   - 4-5: Partial understanding with notable gaps
   - 6-7: Good understanding with minor errors or omissions
   - 8-9: Strong grasp of concepts with minimal errors
   - 10: Complete and perfect answer matching the reference answer

Special Considerations:
- For intervals/ranges: The student's answer must cover the EXACT SAME range as the reference answer
- For multiple solutions: If the reference answer contains multiple solutions (connected by "or"/"and"), all must be present in the student's answer
- For mathematical proofs or procedural questions: Evaluate both the final answer and the method used
- For conceptual questions: Focus on the depth of understanding and clarity of explanation

Your response must always follow this format:
Reasoning: <Provide detailed analysis of the student's answer in relation to the reference answer>
Score: <numerical score between 0 and 10>

The question:
`<PROBLEM>`

The reference answer:
`<REFERENCE_ANSWER>`

The student's answer:
`<STUDENT_ANSWER>`
\end{Verbatim}
\end{tcolorbox}
\caption{Prompt for evaluating model responses against reference answers}
\label{prompt:evaluating_student_answer_with_reference}
\end{figure}

\begin{table}[h]
\centering
\caption{Answer Extraction Patterns}
\begin{tabular}{c|l}
\toprule
\multirow{13}{*}{\shortstack{\textbf{Answer}\\\textbf{Indicators}}}
 & \texttt{The final answer to this question is <ANSWER>} \\
 & \texttt{The correct answer is <ANSWER>} \\
 & \texttt{The best option is <ANSWER>} \\
 & \texttt{The answer is <ANSWER>} \\
 & \texttt{Answer: <ANSWER>} \\
 & \texttt{Answer should be: <ANSWER>} \\
 & \texttt{Answer must be <ANSWER>} \\
 & \texttt{Answer is probably <ANSWER>} \\
 & \texttt{<ANSWER> is correct} \\
 & \texttt{<ANSWER> seems correct} \\
 & \texttt{<ANSWER> is the right answer} \\
 & \texttt{Answer is <ANSWER>} \\
 & ... \\
 
\midrule

\multirow{6}{*}{\shortstack{\textbf{Answer}\\\textbf{Formats}}}
 & \texttt{\textbackslash boxed\{\}} \\
 & \texttt{\textbackslash mathrm\{\}} \\
 & \texttt{\textbackslash mathbf\{\}} \\
 & \texttt{\textbackslash text\{\}} \\
 & \texttt{()} \\
 & \texttt{[]} \\
\bottomrule
\label{tab:answer_extraction}
\end{tabular}
\end{table}



\begin{table}[htbp]
    \caption{Hyperparameters of supervised finetuning.}
    \centering
    \resizebox{\textwidth}{!}{ 
    \begin{tabular}{l|cccccc}
    \toprule
     & \textbf{LR} & \textbf{LR Schedule} & \textbf{Batch Size} & \textbf{Max Length} & \textbf{Warm Up Ratio} & \textbf{Epochs} \\
    \midrule
    SCP-116K & 5e-6 & Cosine & 128 & 16,384 & 0.05 & 3 \\
    NaturalReasoning & 5e-6 & Cosine & 512   & 4,096 & 0.05 & 3 \\
    Nemotron-Science & 5e-6 & Cosine & 128 & 16,384 & 0.05 & 3 \\
    \ourdataset & 5e-6 & Cosine & 512 & 4,096 & 0.05 & 3\\
    \megascience & 5e-6 & Cosine & 512 & 4,096 & 0.05 & 3\\
    \bottomrule
    \end{tabular}
    }
    \label{tab:superparameters_sft}
\end{table}
\end{document}